\documentclass[letterpaper, 10 pt, conference]{IEEEtran}

\IEEEoverridecommandlockouts                            
\usepackage[left=54pt,right=54pt,top=54pt,bottom=54pt]{geometry}

\usepackage{graphicx}
\usepackage{svg}
\usepackage{subcaption}
\usepackage[font=footnotesize]{caption}
\graphicspath{{figures/}{./}}
\usepackage[utf8]{inputenc}
\usepackage[T1]{fontenc}
\usepackage{fancyhdr}    
\usepackage{textcomp}  
\usepackage{amsmath} 
\usepackage{amssymb}
\usepackage{dsfont}
\usepackage{psfrag}
\usepackage{stfloats}
\usepackage{color}
\usepackage{multicol}
\usepackage{algorithmic}
\usepackage{algorithm}
\usepackage{array}
\usepackage{bm}
\usepackage{bbm}
\usepackage{paralist}{}
\usepackage{wasysym}
\usepackage[normalem]{ulem}

\usepackage[bookmarks=true]{hyperref}
\usepackage[hyphenbreaks]{breakurl}

\usepackage{amsthm}

\newtheorem{problem}{Problem}

\newtheorem*{approach*}{ASLaP-HR}

\usepackage{comment}
\usepackage{ulem}
\usepackage{cite}

\usepackage[colorinlistoftodos]{todonotes}

\newcommand{\etal}{\textit{et al. }}

\newcommand{\obj}{\mathcal{I}}

\newcommand{\field}{\phi}
\newcommand{\fieldF}[1]{\field \left( #1 \right)}
\newcommand{\pose}{x}
\newcommand{\poseVec}{\vec{\pose}}
\newcommand{\map}{g}
\newcommand{\mapF}[1]{\map \left( #1 \right)}
\newcommand{\obs}{Z}

\newcommand{\obsSingle}[2][]{\MakeLowercase{\obs}^{(#1)}_{#2}}
\newcommand{\obsVec}[1][]{\left[ \obsSingle[1]{#1}(\pose) , \dots ,  \obsSingle[S]{#1}(\pose) \right]}
\newcommand{\real}[1][]{\mathbb{R}^{#1}}
\newcommand{\dist}[1][]{\mathbb{D}^{#1}}
\newcommand{\interSensor}{h}
\newcommand{\interSensorF}[1]{\interSensor_s \left( #1 \right)}
\newcommand{\learner}{update}
\newcommand{\learnerF}[1]{\learner \left( #1 \right)}

\newcommand{\sumT}{\sum_{t=1}^{T}}

\title{Adaptive Sampling of Latent Phenomena using Heterogeneous Robot Teams (ASLaP-HR)}

\author{Matthew Malencia, Sandeep Manjanna, M. Ani Hsieh, George Pappas, Vijay Kumar
\thanks{We gratefully acknowledge the support from ARL Grant DCIST CRA W911NF-17-2-0181, NSF Grant CNS-1521617,
ARO Grant W911NF-13-1-0350, ONR Grants N00014-20-1-2822 and
ONR grant N00014-20-S-B001, and
Qualcomm Research. The first author acknowledges support from the National Science Foundation Graduate Research Fellowship under Grant No. DGE-1845298.} 
\thanks{The authors are with GRASP Laboratory, Levine Hall 4th floor, University of Pennsylvania, 3330 Walnut Street, Philadelphia, PA 19104-6228. \textit{corresponding authors emails:}
{\tt\footnotesize \{malencia,msandeep\}@seas.upenn.edu}}%
}

\newgeometry{left=54pt,right=54pt,top=72pt,bottom=54pt}
\begin{document}
\maketitle


\begin{abstract}
In this paper, we present an online adaptive planning strategy for a team of robots with heterogeneous sensors to sample from a latent spatial field using a learned model for decision making. Current robotic sampling methods seek to gather information about an observable spatial field. However, many applications, such as environmental monitoring and precision agriculture, involve phenomena that are not directly observable or are costly to measure, called latent phenomena. 
In our approach, we seek to reason about the latent phenomenon in real-time by effectively sampling the observable spatial fields using a team of robots with heterogeneous sensors, where each robot has a distinct sensor to measure a different observable field.
The information gain is estimated using a learned model that maps from the observable spatial fields to the latent phenomenon. This model captures aleatoric uncertainty in the relationship to allow for information theoretic measures. 
Additionally, we explicitly consider the correlations \textit{among} the observable spatial fields, capturing the relationship between sensor types whose observations are not independent. We show it is possible to learn these correlations, and investigate the impact of the learned correlation models on the performance of our sampling approach. Through our qualitative and quantitative results, we illustrate that empirically learned correlations improve the overall sampling efficiency of the team. We simulate our approach using a data set of sensor measurements collected on Lac Hertel, in Quebec, which we make publicly available.

\end{abstract}


\section{Introduction} \label{Sec:Intro}
Robot teams have proven to be powerful tools for environmental monitoring, drastically improving efficiency over fixed sensor networks or manual sampling \cite{dunbabin_robots_2012}. This effectiveness is driven in large part due to their ability to move and collect samples from multiple locations, and due to the sampling strategies that decide where measurements should be taken \cite{manjanna_scalable_2021}.
However, many applications involve \textbf{phenomenon that either cannot be directly measured by sensors or require slow and costly sensors to measure, called \textit{latent phenomena} or \textit{latent spatial fields}}. For example, in agricultural applications, chlorophyll is measured as a proxy for health or as an indicator of an infestation \cite{barker_relationships_1976}. And in marine settings, scientists infer aquatic health by measuring various chemical concentration indicators \cite{dunbabin_robots_2012}. In this paper, we answer the question: \textbf{how can robot teams choose sensing locations that best inform a latent phenomenon?} We formulate this question as an Adaptive Sampling of Latent Phenomena (ASLaP) problem and propose an informed path planning technique for a heterogeneous team of robots (ASLaP-HR) that seeks to estimate a quasi-static latent spatial field. We assume that the observable and latent spatial fields change slowly or remain static at least for the time span of a robotic survey and that robot team's communication is not constrained with the region.
 
\begin{figure}
  \begin{subfigure}{0.98\columnwidth}
    \centering
    \includegraphics[width = 0.3\columnwidth]{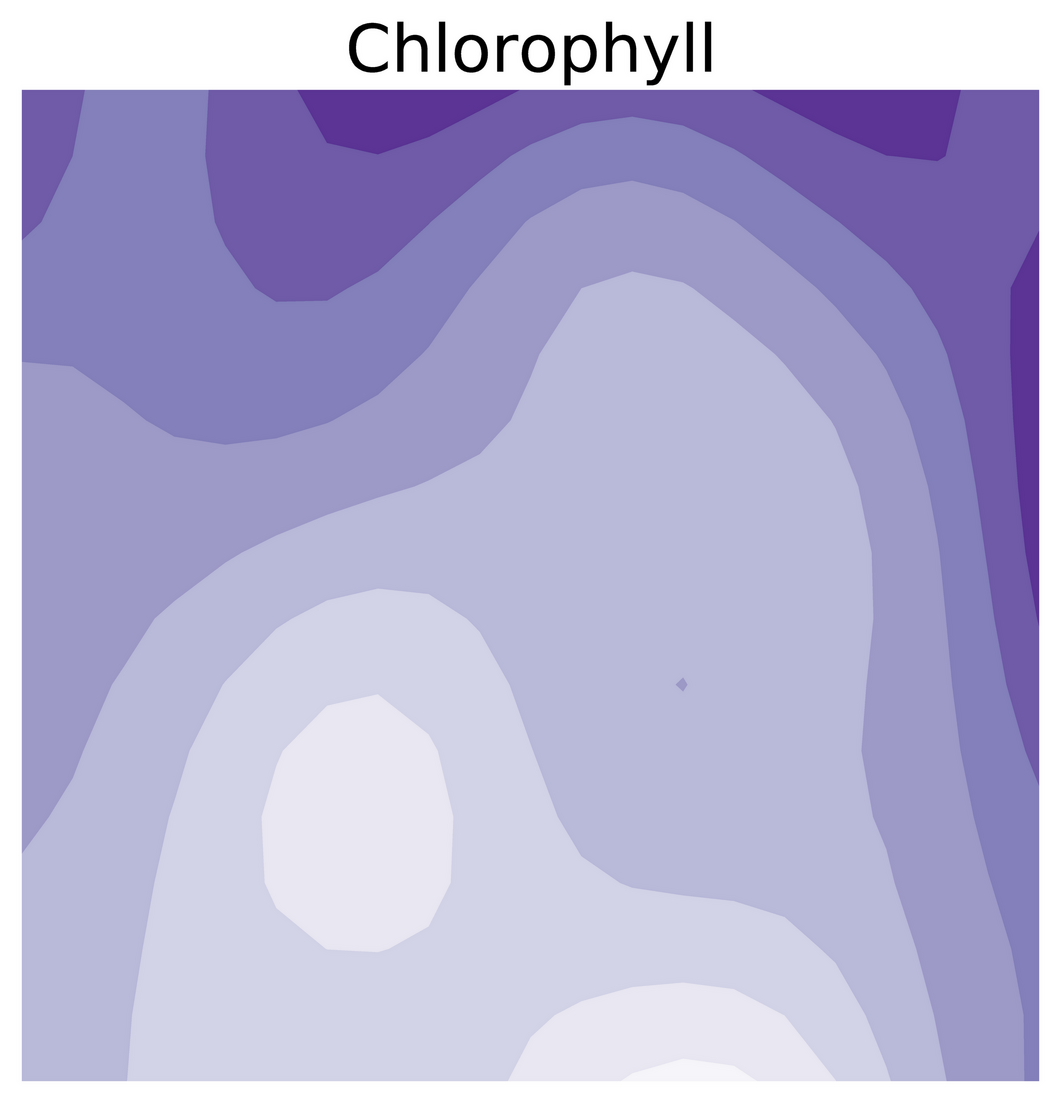}
    \includegraphics[width = 0.3\columnwidth]{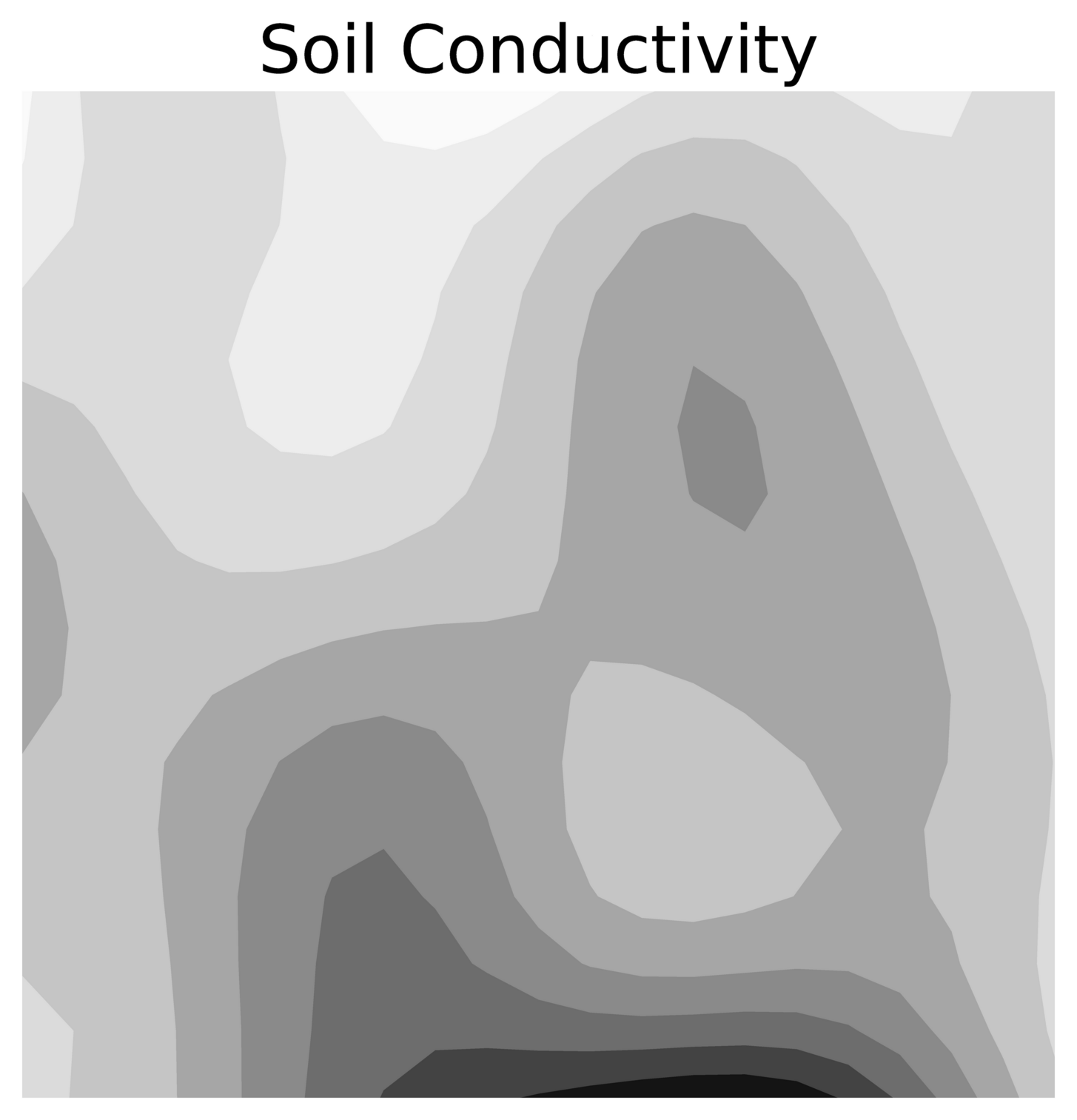}
    \includegraphics[width = 0.3\columnwidth]{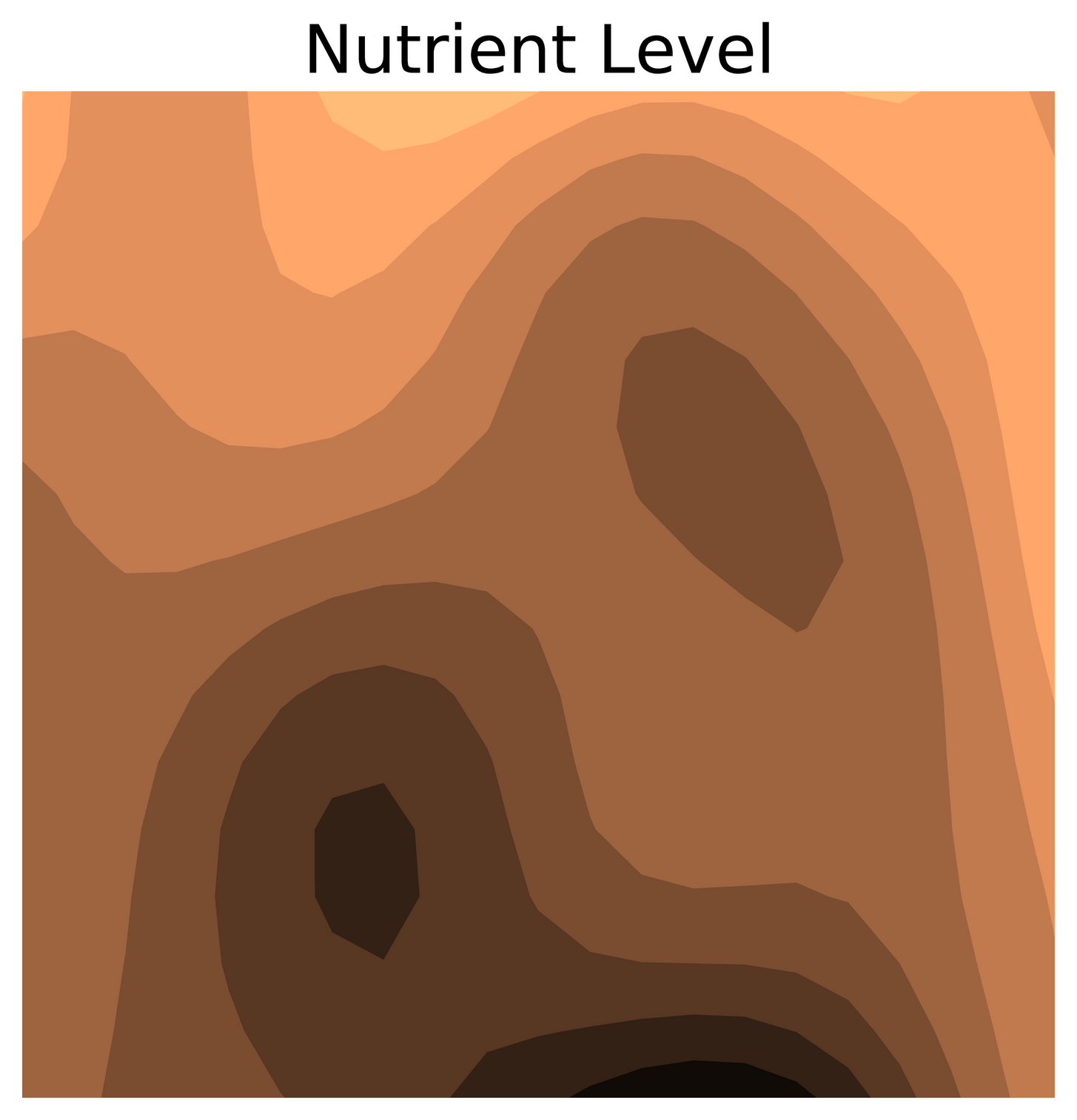}
    \caption{Spatial fields.} \label{fig:example-fields}
  \end{subfigure}

  \begin{subfigure}{0.44\columnwidth}
    \centering
    \includegraphics[width = \columnwidth]{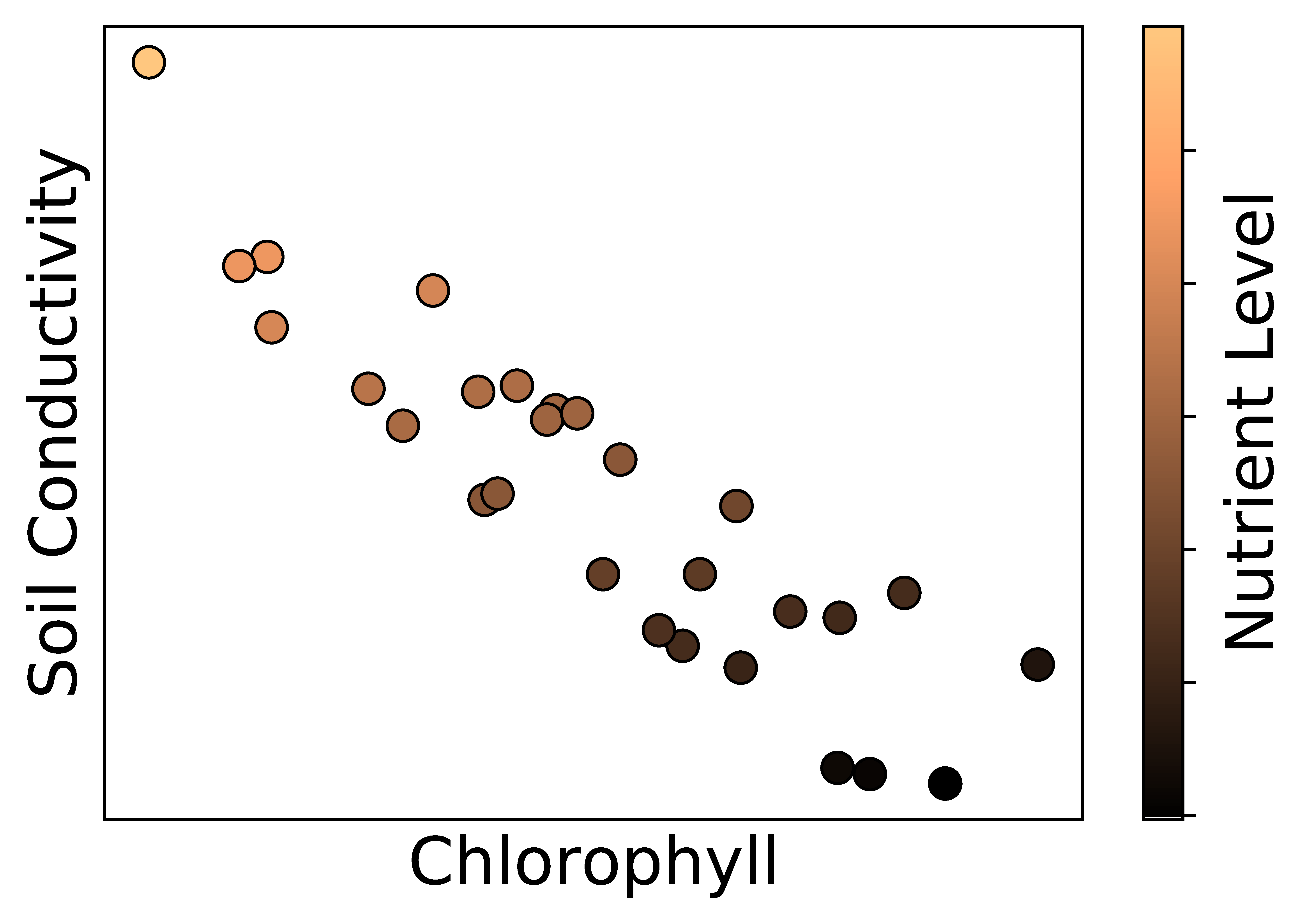}
    \caption{Historical Data} \label{fig:example-data}
  \end{subfigure}
  \begin{subfigure}{0.47\columnwidth}
    \centering
    \includegraphics[width = \columnwidth]{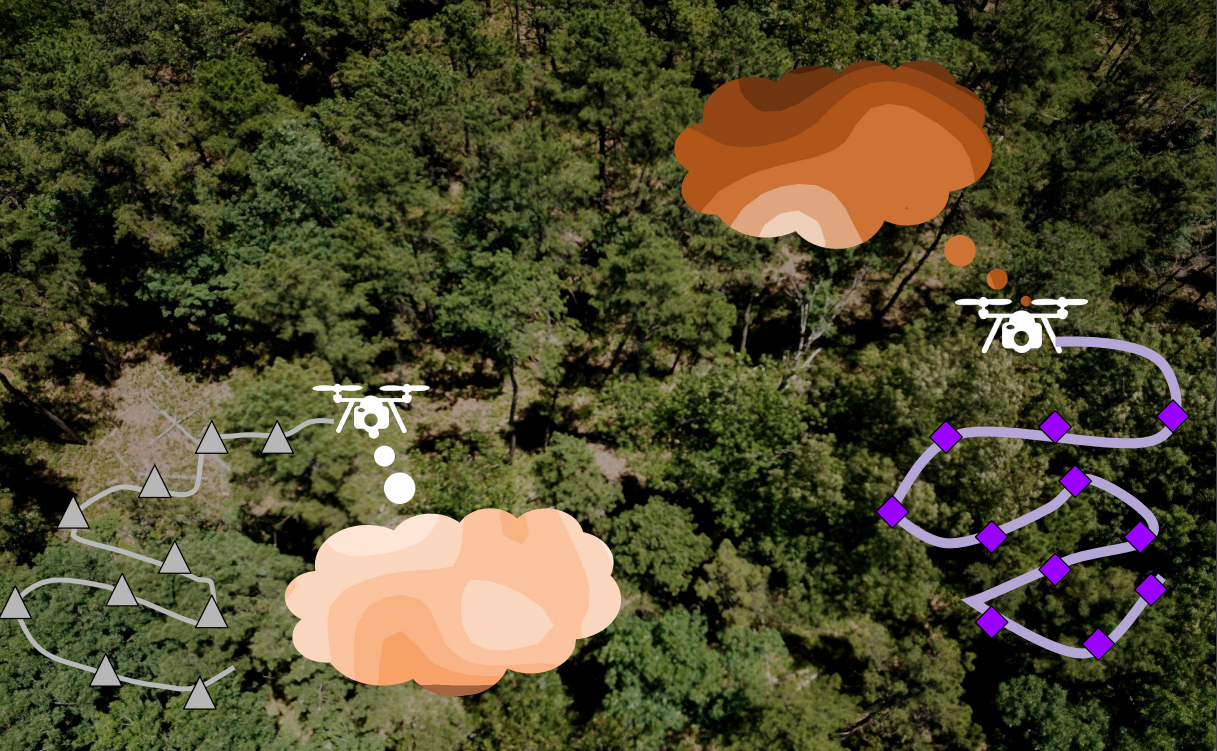}
    \caption{Multi-robot information gathering.} \label{fig:example-fields}
  \end{subfigure}%
    \caption{An example scenario: \textit{(a)} Farms can be described by various spatial fields. A spatial map of nutrient levels is most informative to a farmer, but it requires costly and time-consuming sampling and analysis of leaves. Chlorophyll and soil conductivity spatial fields can be measured quickly and cheaply by in-situ sensors. \textit{(b)} Historical data is used to train a model to predict nutrient levels using chlorophyll and soil conductivity measurements. \textit{(c)} A team of robots can measure chlorophyll and soil conductivity relatively efficiently and use the learned model from (b) to make real-time decisions about where and what to sample next to get best estimates of nutrition.}
    \label{fig:example}\vspace{-1.5em}
\end{figure}

Consider the example scenario presented in Fig.~\ref{fig:example}: A farmer monitors plant nutrition by sending leaf samples to a lab for analysis. However, cheap and fast sensors, such as chlorophyll and soil conductivity sensors, can provide in-situ proxy measurements that estimate plant nutrition. Our approach to this scenario uses historical data of co-located chlorophyll, soil conductivity, and analyzed leaf samples to learn the relationship between the sensor measurements and nutrient levels. Then, a team of robots (equipped with soil conductivity and chlorophyll sensors) collaboratively estimates the nutrition levels of the whole farm (or plot) by using the learned model to inform where and which sensor measurements will provide the best estimate of nutrition.

Current work on sampling approaches for robot teams either optimize objectives on observable spatial fields  \cite{quattrini_li_multi-robot_2020, almadhoun_survey_2019} or assume an exact proxy for a latent spatial field \cite{manjanna_heterogeneous_2018}. \restoregeometry However, this does not necessarily align with optimizing relevant information of the latent phenomenon of interest \cite{arora_multi-modal_2019}. After collecting data of their observable spatial fields, these teams rely on human experts for higher level decision making about the phenomenon of interest. 

Heterogeneity in robot teams has been considered in robotic sampling \cite{salam_heterogeneous_2021}, with a focus on diverse robot types, e.g., aerial vs ground robots \cite{manderson_heterogeneous_2019}. While some papers consider sensor heterogeneity, this area is limited to sensors that align with prescribed roles \cite{manjanna_heterogeneous_2018} and sensors that are homogeneous in type but heterogeneous in cost, range, and accuracy \cite{shi_adaptive_2020}. In this work, we investigate \textbf{heterogeneity in sensor types, where each type measures a different spatial field}. These diverse observations can combat uncertainty of the latent phenomenon because real world sensor spaces are not independent and map nonuniformly to the latent spatial field.

In order to leverage the strength of heterogeneous sensors, we use data-driven methods to model the relationship between the observable spatial fields and the latent phenomenon. 
Most machine learning methods are limited by the resolution of the input data, e.g., using satellite data to make predictions about entire plots of land rather than individual locations or crops \cite{chlingaryan_machine_2018}, or they postprocess spatial data to gain meaningful insights. Our work seeks to enable predictions that reason directly about the latent spatial field in real time while also providing the precision of studying localized phenomena.

\subsection{Related Work.}\label{Sec:Prelim}
There are myriad problem spaces that involve path planning for mobile robots. Bayesian optimization approaches seek to find the maximum of a spatial field \cite{marchant_sequential_2014, oliveira_bayesian_2019,  tan_gaussian_2018}; exploration and active mapping optimize the observation of an unknown area \cite{yamauchi_frontier-based_1998, lluvia_active_2021, popovic_informative_2020}; and active information gathering or target tracking focus on estimating the state of dynamic hidden states \cite{tzes_technical_2021, kantaros_sampling-based_2021,salam_adaptive_2019}. The objective functions studied among these applications can be categorized as information-theoretic driven \cite{fung_coordinating_2019, bai_information-driven_2021}, such as mutual information \cite{low_information-theoretic_2009, singh_efficient_2009}, and sample driven, as is the case of hotspot sampling \cite{low_adaptive_2008, manjanna_policy_2018}. This work seeks to maximize the expected information gained by a team of heterogeneous robots.

Heterogeneity of multi-robot teams can vary drastically. Manjanna~\etal use two autonomous surface vehicles, one with a water quality sensor and the other with a water sampler, where the sensor vehicle explores to inform the sampler's decisions \cite{manjanna_heterogeneous_2018}. Similarly, Manderson~\etal use an aerial vehicle to inform the decisions of a ground vehicle \cite{manderson_heterogeneous_2019}. Chand and Carnegie investigated heterogeneous processing and sensing capabilities to assign tasks to different robot types, where robots with low-value sensing do computation rather than exploration tasks \cite{chand_mapping_2013}. Shi~\etal use a Voronoi approach where the space is partitioned according to heterogeneous capabilities, like mobility constraints in ground versus aerial vehicles \cite{shi_adaptive_2020}. Similarly, Cai~\etal consider heterogeneous capabilities such as field of view and traversability of different environment types \cite{cai_non-monotone_2021}. Others still consider resource constraints such as energy budgets and fuel limitations \cite{newaz_multi-robot_2021, notomista_multi-robot_2022, arora_randomized_2017}. Whereas Chand and Carnegie model different levels of sensing ability (e.g., low vs high), we model qualitatively different sensor types, which all take high quality sensor measurements in their respective observable space. Manjanna~\etal and Manderson~\etal investigate different robot types (sensor vs sampler, and aerial vs ground sensing, respectively), but their investigations focus on two prescribed roles. In this work, we do \textit{not} consider heterogeneity in robot resources (e.g., computation power, or mobility type) but rather heterogeneity in sensor types; we model an arbitrary number of sensor types where tasks and roles are not pre-designed.

Recent work has only just begun to develop models and algorithms to make real-time decisions with respect to a latent phenomenon of interest. Arora~\etal study multi-modal active perception, where measurements must be combined with domain knowledge to deduce information about variables or phenomena that cannot be directly measured \cite{arora_multi-modal_2019}. Their paper investigates a single robot with multiple sensors that make discrete observations and maps the observations to the variables of interest through an expert-designed Bayesian network. In this work, we consider a multi-robot team where sensor measurements are continuous and we learn a continuous representation mapping from the observations to the latent phenomenon of interest. 

\subsection{Contributions} We formalize adaptive informative sampling for multiple continuous sensors to maximize information theoretic objective of a latent phenomenon. We use a robot team with heterogeneous sensor types to measure observable spatial fields. A learned model from historical data enables the robot team to infer about the latent spatial field in real time and use these inferences to choose sample locations that maximize the information gain of the latent phenomenon. Additionally, we learn correlations \textbf{among} the multiple observable spatial fields; these correlations are learned offline and are used to make real-time sampling location decisions. We investigate the impact of these learned correlations on the adaptive sampling of the latent field. Our main contributions are:
\begin{itemize}
    \item We formulate the ASLaP problem as the maximization of the expected information of a latent spatial field. 
    \item We develop ASLaP-HR, an adaptive sampling approach for a team of robots with heterogeneous sensors using a learned mapping from observations to the latent phenomenon.
    \item We show that the learned correlations among the observable spatial fields improve the sampling efficiency of our adaptive sampling approach.
\end{itemize} 

In Section~\ref{Sec:Problem}, we present a novel ASLaP problem formulation that is not restricted to myopic or non-myopic approaches, nor does it restrict motion uncertainties or arbitrary sets of actions,\footnote{The problem formulation considers a discrete action set. Using continuous dynamics would require slight reformulation.} though the methods presented in Section~\ref{Sec:Approach} are myopic and do not consider motion uncertainty. We validate our approach by running experiments on real sensor data collected on a lake. The sensor data, which we make publicly available \footnote{This data is available at: \url{https://github.com/SandeepManjanna/water_quality_dataset}}, includes co-located measurements of ten water quality parameters. Our experiments use three of the ten sensor fields, two as the sensor spaces and one as the latent variable of interest. The experimental results show that the learned correlations among the observable spatial fields improve overall sampling. 
Future work will investigate the online learning of these correlations.


\section{Problem Definition} \label{Sec:Problem}
Consider a region of locations $\pose \in \real[2]$ with a spatial field $\field:~\pose\mapsto\real$, such as temperature.\footnote{We study 2-dimensional spatial fields in this work, though 3-dimensional space could be considered.} Adaptive informative sampling seeks to estimate this spatial field $\field$. The estimate is a belief model, $\Hat{\field}: \pose\mapsto\dist[]$, where $\dist[]$ is the set of  distributions in $\real[]$; for each $\pose$, $\Hat{\field}(\pose)$ is a one-dimensional distribution that estimates $\fieldF{\pose}$. To model $\Hat{\field}$, robots make observations at locations $\pose$ that maximizes information gain, $\obj$. Often, $\obj$ is calculated using Shannon entropy, $H$. The information gain with respect to $\field$ of a measurement at $\pose$ is the change in entropy of the field estimate due to the new measurement: $\obj_{\hat{\field}}(x)= H(\Hat{\field}) - H(\Hat{\field} | \pose)$. Additionally, adaptive sampling problems often include resource constraints, such as energy budgets \cite{tiwari_multi-robot_2019} or sampling budgets \cite{manjanna_heterogeneous_2018}. We limit the number of measurements using a time horizon constraint, $T$.\footnote{The constraint $T$ can be interpreted as an energy budget, assuming each action takes equal energy, or as a sensing budget of at most $T$ measurements per robot (we assume one measurement is taken each time step). Other works define a cost per action/measurement and constrain the sum of costs.}
Therefore, a robot seeks a series of measurements $\pose_1, \dots, \pose_T$ that maximize $\sumT \obj_{\hat{\field}}(x_t)$. This extends to a team of $N$ robots where the vector $\poseVec_t = (\pose^{(1)}_t, \dots, \pose^{(N)}_T)$. This type of problem formulation is found in the previous literature on informative path planning \cite{arora_multi-modal_2019, chen_multi-objective_2019, hollinger_sampling-based_2014, low_adaptive_2008}. Conversely, we consider a latent spatial field where \textbf{$\field$ is not directly observeable}, yielding the Adaptive Sampling of Latent Phenomena (ASLaP) problem formulation:

\begin{problem}\label{prob:base}
\emph{ASLaP:}
Given a team of $N$ agents, find observation locations $\mathbf{x}_t$ at all time steps in time horizon $T$ that maximize the sum of information gained relative to the latent spatial field $\phi$. Formally:
\begin{equation}\label{eq:base}
\underset{\poseVec_1, \dots, \poseVec_T}{\text{arg\,max}}  \sumT \obj_{\hat{\field}} \left[ \poseVec_t \right]
\end{equation}
\end{problem}

The novel problem formulation in Problem~\ref{prob:base} is challenging because \textbf{$\field$ is a latent phenomenon}. This breaks the fundamental assumption that the robots can directly measure the field $\field$, thus requiring new methods that enable inference over this latent spatial field. In addition to this novel problem formulation, this work is the first to take advantage of learned spatial field correlation models to improve adaptive sampling.

In the following section, we present an approach using a team of robots with heterogeneous sensors, where the different sensor types measure observable spatial fields to collectively inform the estimate of the latent phenomenon by using a mapping that is learned from historical data.
Additionally, we investigate how learned correlation models among these observable spatial fields can be leveraged to improve the efficiency of this team's sampling.

\section{Approach} \label{Sec:Approach}
\begin{figure}[tb]
    \centering
    \includegraphics[width = 0.25\columnwidth]{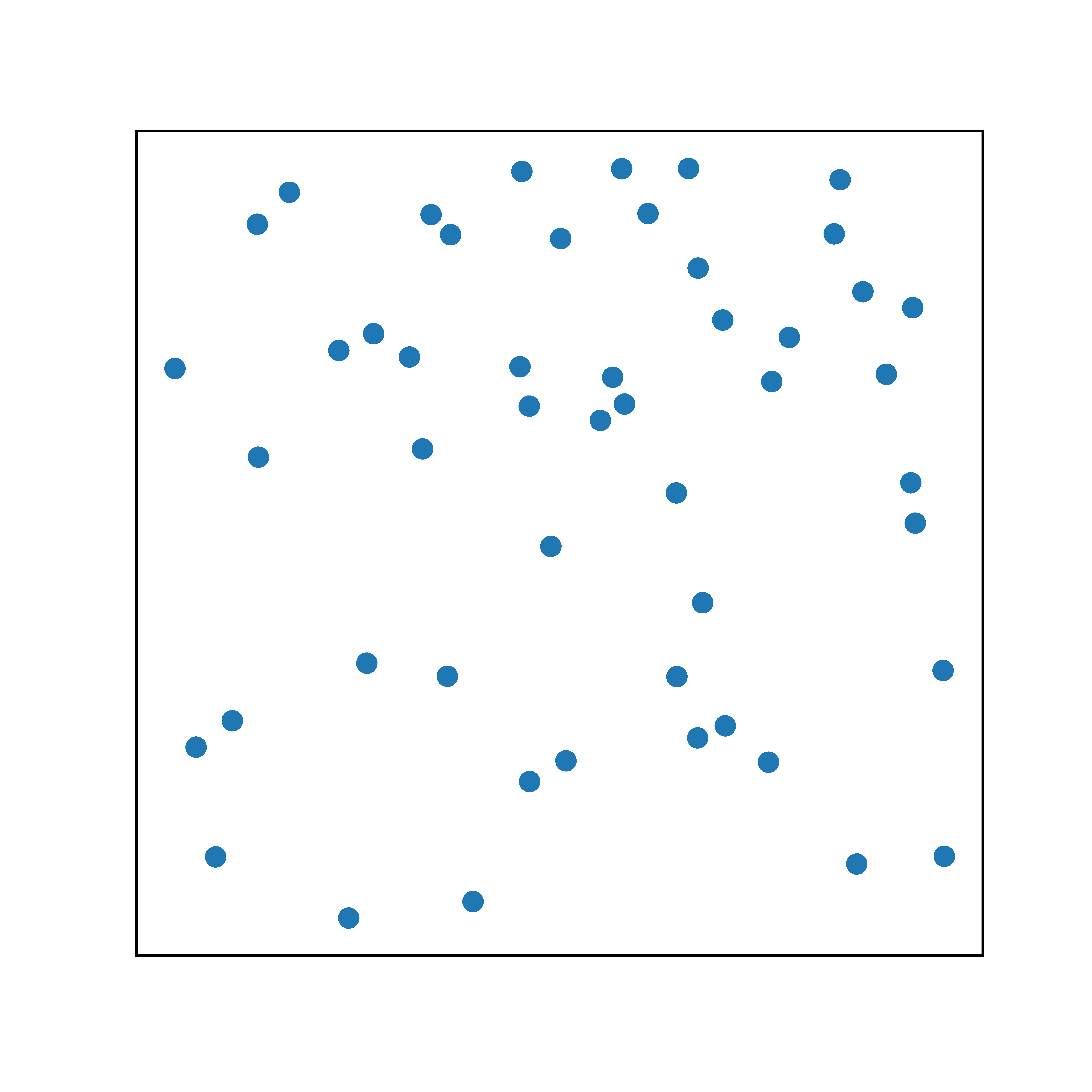}
    \includegraphics[width = 0.385\columnwidth]{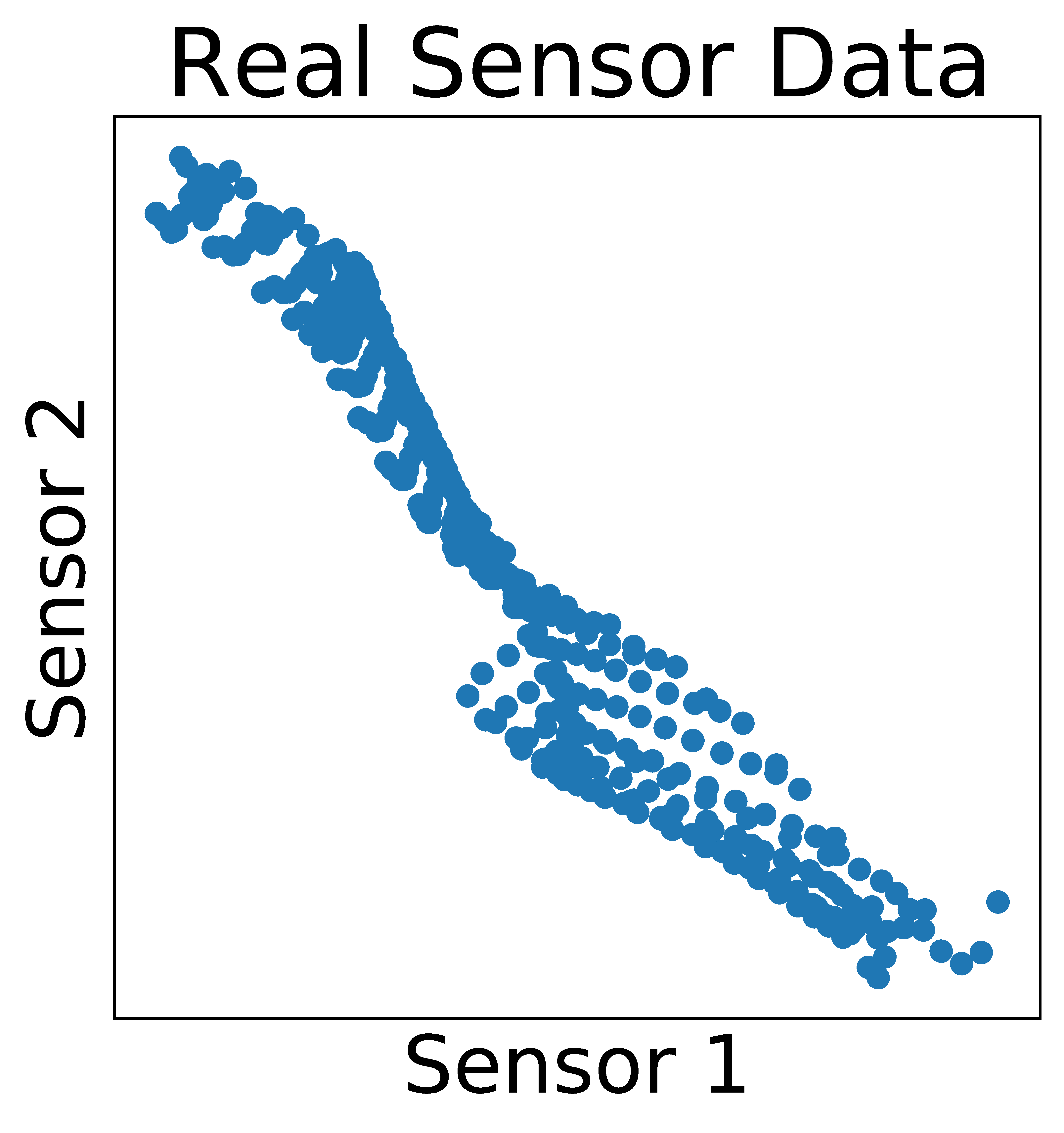}
    \includegraphics[width = 0.25\columnwidth]{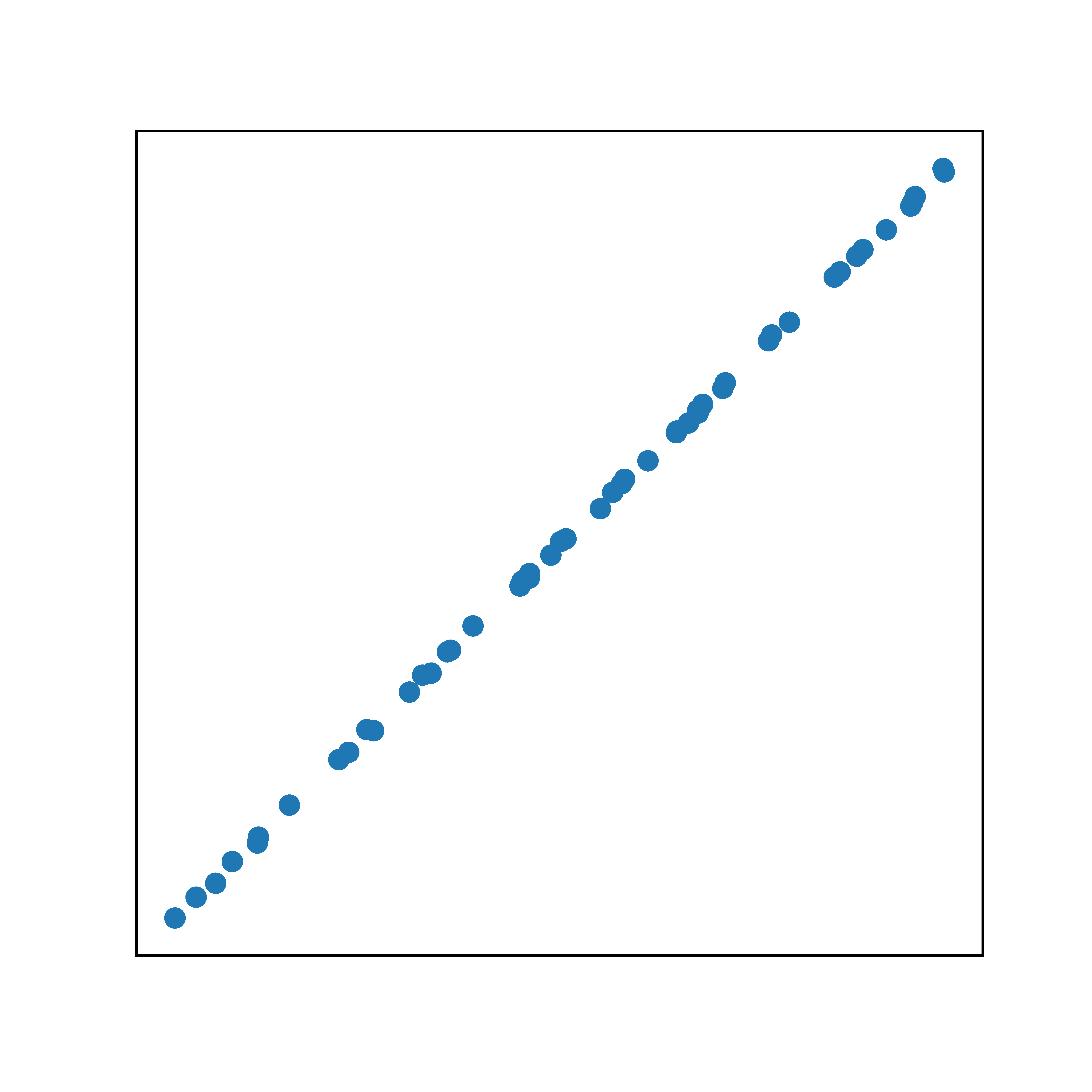}
    \includegraphics[width = 0.875\columnwidth]{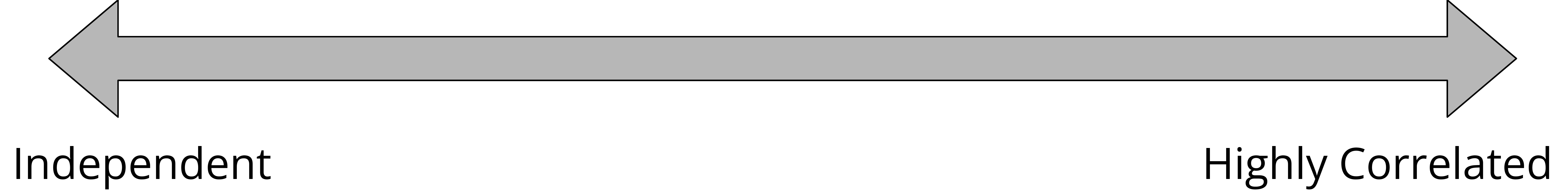}
    \caption{The mutual information between different sensor types. Independent sensor types are independent sources of information. For perfectly correlated sensor types, co-located measurements from a second sensor type provide zero additional information. In reality, different sensor types (and the respective measurable spatial fields), have complex dependencies with mutual information that is complex, unknown function of the measurements. The middle scatter plot above shows the relationship of real sensor data that is used in our experiments in Section~\ref{Sec:Results}.}
    \label{fig:intersensor} \vspace{-1.5em}
\end{figure}

We consider a heterogeneous team of $N$ robots with $S$ sensor types among the team, where $S=N$, which means that each robot has a distinct sensor type. It is possible to extend our approach to an arbitrary number of robots sensor combinations where $S\neq N$.\footnote{For $S\neq N$, there are $N^{(s)}$ robots per sensor type such that $\sum_{s=1}^{S} N^{(s)}=N$.}
Each sensor type measures a spatial field, and, as with real world observeable spatial fields, sensor measurements from different sensor types are \textit{not} independent of one another, as seen in Fig.~\ref{fig:intersensor}. We use the index $s = 1, \dots , S$ for both robots and sensors. Each robot $r_s$ can measure a spatial field $\obsSingle[s]{}:\pose\mapsto\real[]$.  Let $\pose^{(s)}_t$ be the location of robot $r_s$ at time $t$ and let the respective observation be $\obsSingle[s]{} ( \pose^{(s)}_t )$. For ease of notation, the observation made by robot $r_s$ at time $t$ will be denoted by $\obsSingle[s]{t}(\pose)$. Let $\obs:\pose\mapsto\real[S]$ be the $S$-dimensional observable spatial field, where $\obs(\poseVec)= \obsVec$. The spatial field $\obs$ is unknown but observable, therefore, online measurements are used to maintain a belief estimate $\Hat{\obs}:\pose\mapsto\dist[S]$.

With this team of heterogeneous multisensor robots, we seek to build an estimate of the latent spatial field. To do this, a mapping $\map$ from observations in $\obs$ to the latent spatial field $\field$ is learned from historical data. Robots make measurements of $\obs$, build an estimate of the observable spatial field $\Hat{\obs}$, and then use the learned mapping $\map:\Hat{\obs}\mapsto\Hat{\field}$ to maintain the estimate of the latent spatial field $\Hat{\field}$. Learning $g$ is challenging because the mapping from $\Hat{\obs}\in\dist[S]$ to $\Hat{\field}\in\dist[]$ has aleatoric uncertainty, meaning even an exact measurement of an observable spatial field yields an uncertain latent spatial field value. Multiple machine learning methods can capture this aleatoric uncertainty \cite{hullermeier_aleatoric_2021}, such as the Gaussian Process approach that we use in Section~\ref{Sec:Results}. Because the goal of our approach is to create a good estimate of the latent phenomenon, the observation locations $\pose$ are chosen to maximize the information gain with respect to $\field$. Using the learned mapping $g$, the information gain is now calculated by $\obj_{\map(\Hat{\obs})}(x)= H(\map(\Hat{\obs})) - H(g(\Hat{\obs} | \pose))$, which yields the Adaptive Sampling of Latent Phenomena using Heterogeneous Robots (ASLaP-HR) approach:

\begin{approach*}\label{prob:adaptive_latent}
Given a team of $S$ robots, each equipped with a distinct sensor $s$, each robot $s$ takes measurements of the field $\obsSingle[s]{}$ to collectively form an estimate $\Hat{Z}$ of the entire observable spatial field. This in turn allows estimates of the latent spatial field by using $g$, a mapping from the observable spatial fields to the latent spatial field that must be learned from historic data. At each time step $t$, choose the observation location for each robot, $\poseVec_t = (x^{(1)}_t, \dots, x^{(N)}_T)$, that maximize the information gain $\obj_{\map(\hat{\obs})}$ with respect to this latent phenomenon. Formally:
\begin{equation}\label{eq:adaptive_latent}
\underset{\poseVec_1, \dots, \poseVec_T}{\text{arg\,max}}  \sumT \obj_{\map(\hat{\obs})}\left[ \poseVec_t \right]
\end{equation}
\end{approach*}

While prior works reason about observable spatial fields, ASLaP-HR reasons about a latent phenomenon through $g$. The following section presents an ASLaP-HR planning algorithm, which leverages a learned correlation among the observable spatial fields for improved information gathering. Section~\ref{Sec:Results} presents a specific implementation of this algorithm and shows the improvement gained by leveraging the learned correlation.

\subsection{Planning Algorithm Details}

We present Algorithm~\ref{alg:planner} for ASLaP-HR, where a team of $S$ robots take measurements of $\obs$, build an estimate of the observable spatial field $\Hat{\obs}$, and then use the learned mapping $\map$ to build an estimate of the latent spatial field $\Hat{\field}$. Algorithm~\ref{alg:planner} takes the following inputs:
\begin{itemize}
    \item {\makebox[2.25cm]{$\poseVec_0 \in \real[2\text{x}S]$\hfill} initial positions of all robots}
    \item {\makebox[2.25cm]{$T \in \mathbb{Z}^{+}$\hfill} time horizon}
    \item {\makebox[2.25cm]{$\map:\Hat{\obs}\mapsto\Hat{\field}$\hfill} mapping from observations to latent field}
    \item {\makebox[2.25cm]{$h_s:\obsSingle[s]{}\mapsto\obs$\hfill} correlation model for sensor type $s$}
\end{itemize}

Algorithm~\ref{alg:planner} begins with each robot $r_s$ observing the spatial field $\obsSingle[s]{}$ at their initial position $\pose^{(s)}_{t}$, yielding measurements $\obs(\poseVec_0)$. These measurements are used to update the belief $\Hat{\obs}$ (line~\ref{alg:line:learner_init}) which is then used by $\map$ in line~\ref{alg:line:map_init} to update the belief $\Hat{\field}$. In Section~\ref{Sec:Results}, we model $\Hat{\obs}$ using Gaussian Processes, which produce inputs to the learned model $g$, which in turn outputs $\Hat{\field}$.
The team then communicates the sensor measurements.\footnote{We assume robots can communicate with each other within the region, and that time needed for communication and belief updating is significantly shorter than the time-step duration.} 
For each time step $t$, and robot $r_s$, the next waypoint $\pose_t^{(s)}$ is selected to maximize the expected information gain. Calculating the information gain using Shannon Entropy (as shown in Section~\ref{Sec:Problem}) is computationally intractable, therefore, the variance of $\Hat{\field}$ is used as an estimate of the expected information gain \cite{krause_near-optimal_2008}.

Next, $sense(s, \pose_t^{(s)})$ on line~\ref{alg:line:move} returns a new sensor measurement $\obsSingle[s]{t}$.
This new sensor measurement is used to refine the model $\interSensor_s$ using any measurements from other sensors that are located together at $\pose^{(s)}_{t}$. We assume that GPS localizes the robots; the uncertainty in this localization does not hinder the estimate of the beliefs. The new sensor measurements are communicated, and the correlation models $\interSensor_s$ are used in line~\ref{alg:line:infer} to update the beliefs $\Hat{\obs}$ at the points $\pose_t^{(s)}$. Line~\ref{alg:line:update} updates the belief $\Hat{\obs}$, and line~\ref{alg:line:map_update} updates the belief $\Hat{\field}$. Algorithm~\ref{alg:planner} then repeats this sequence to iterate over all $T$ time steps, at the end returning the most up-to-date belief $\Hat{\field}$ of the latent phenomenon. The next section elaborates on the line~\ref{alg:line:infer} update of $\Hat{\obs}$.

\begin{algorithm}
\caption{Pseudocode for ASLaP-HR Planner}
\begin{algorithmic}[1] \label{alg:planner}
    \REQUIRE $\poseVec_0, T, g, h_1, \dots, h_S$
    \ENSURE Estimated latent spatial field $\Hat{\field}$
    \STATE Initialize $\Hat{\field}, \Hat{\obs}$
    \STATE Measure $\obs(\poseVec_0)$ \label{alg:line:first_measurement}
    \STATE $\Hat{\obs} \xleftarrow{} \learnerF{\obs(\poseVec_0)}$ \label{alg:line:learner_init}
    \STATE $\Hat{\field} \xleftarrow{} \mapF{\Hat{\obs}}$ \label{alg:line:map_init}
    \FOR{$t = 1, \dots , T$}
        \FOR{$s = 1, \dots , S$}
            \STATE $\pose^{(s)}_{t} \xleftarrow{} \underset{\pose}{arg\,max} \quad  \obj_{\map(\hat{\obs})} \left[ \poseVec_t \right]$\label{alg:line:acquisition}
            \STATE $\obsSingle[s]{t} =  sense(s, \pose^{(s)}_{t})$ \label{alg:line:move}
            \STATE $\interSensor_s \xleftarrow{} refine(\pose^{(s)}_{t}, \obsSingle[s]{t})$ \label{alg:line:udate_intersensor}
        \ENDFOR
        \STATE $\Hat{\obs}(\pose^{(s)}_{t}) = \interSensorF{\obsSingle[s]{t}}  \; \forall \; s = 1, \dots , S$ \label{alg:line:infer}
        \STATE $\Hat{\obs} \xleftarrow{} \learnerF{\obs(\poseVec_t)}$ \label{alg:line:update}
        \STATE $\Hat{\field} \xleftarrow{} \mapF{\Hat{\obs}}$ \label{alg:line:map_update}
    \ENDFOR
    \RETURN 
\end{algorithmic}
\end{algorithm}


\subsection{Learned Correlations among the Observable Spatial Fields} \label{sec:interSensor}

Through this point in Section~\ref{Sec:Approach}, we have presented the basic structure of Algorithm~\ref{alg:planner}. The last core piece, presented here and investigated in Section~\ref{Sec:Results}, is the learned correlations among the observable spatial fields $\interSensor_s$. Algorithm~\ref{alg:planner} does not require line~\ref{alg:line:infer}. However, we hypothesize that \textbf{heterogeneous robots can leverage the relationships between sensor-types to achieve more efficient information gathering.} 

When $\map$ updates the belief of the latent spatial field at a point $\pose$, it calculates $\Hat{\field}(\pose)$ as a function of the entire observable field $\Hat{\obs}(\pose)$. Therefore, the precision of the value and variance of $\Hat{\field}(x)$ depends on the precision of the entire belief $\Hat{\obs}$. However, a robot $r_s$ measures only part of $\Hat{\obs}$, specifically $\obsSingle[s]{}(\pose)$. 

The precision of the variance of $\Hat{\field}(x)$ is especially important because it informs the selection sensing locations. If $\Hat{\field}(\pose)$ \textit{underestimates} variance, line \ref{alg:line:acquisition} calculates a lower expected information gain than is actually present, causing robots to avoid informative measurements. If $\Hat{\field}(\pose)$ \textit{overestimates} variance, robots may take redundant measurements because they expect more information than is actually available.

In Algorithm~\ref{alg:planner}, we see that the observable spatial field correlation models $\interSensor_s$ are used to improve the estimate of $\Hat{\obs}$, which in turn improves the estimate of $\Hat{\field}$. Given a measurement $\obsSingle[s]{}(\pose)$, the correlation model $\interSensor_s$ estimates the value (with variance) of the entire observable spatial field $\Hat{\obs}(\pose)$. These models capture the aleatoric uncertainty in the correlations among the observable spatial fields. For example, if there is low aleatoric uncertainty between two sensor types, one sensor acts almost as a noisy sensor of another spatial field. Conversely, correlations with high aleatoric uncertainty only marginally diminish the variance of other spatial fields.
The line~\ref{alg:line:infer} of Algorithm~\ref{alg:planner} shows how the correlation models are used after a robot moves to a new $\pose$ and takes a new sample $\obsSingle[s]{}$. The new sample is the input to model $\interSensor_s$ that provides updated estimates of the value and variance of the entire observable space at that location, $\Hat{\obs}(x)$.

\section{Results} \label{Sec:Results}
\begin{figure}[tb]
    \centering
    \includegraphics[width = 0.33\columnwidth]{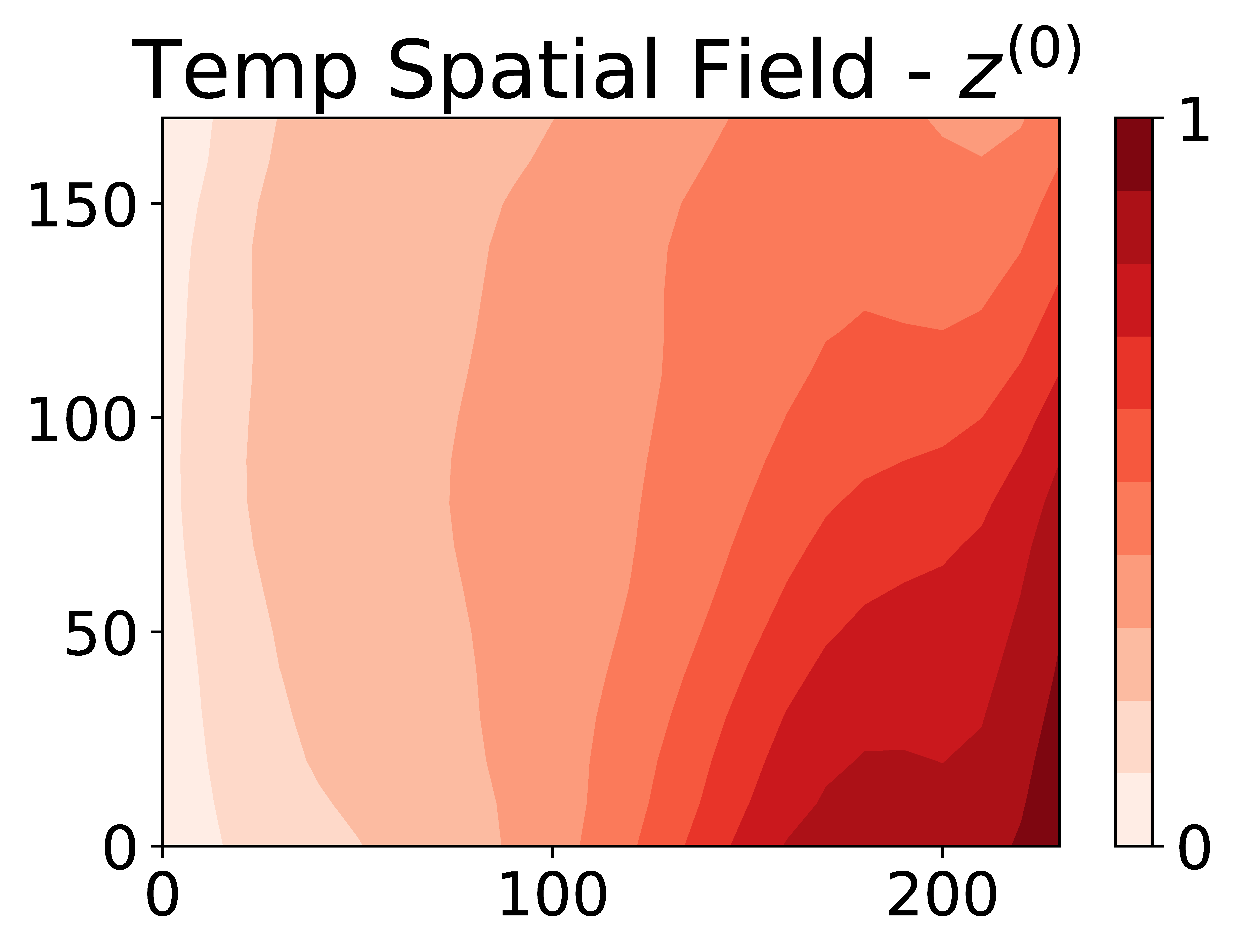}\hspace{-0.8em}
    \includegraphics[width = 0.33\columnwidth]{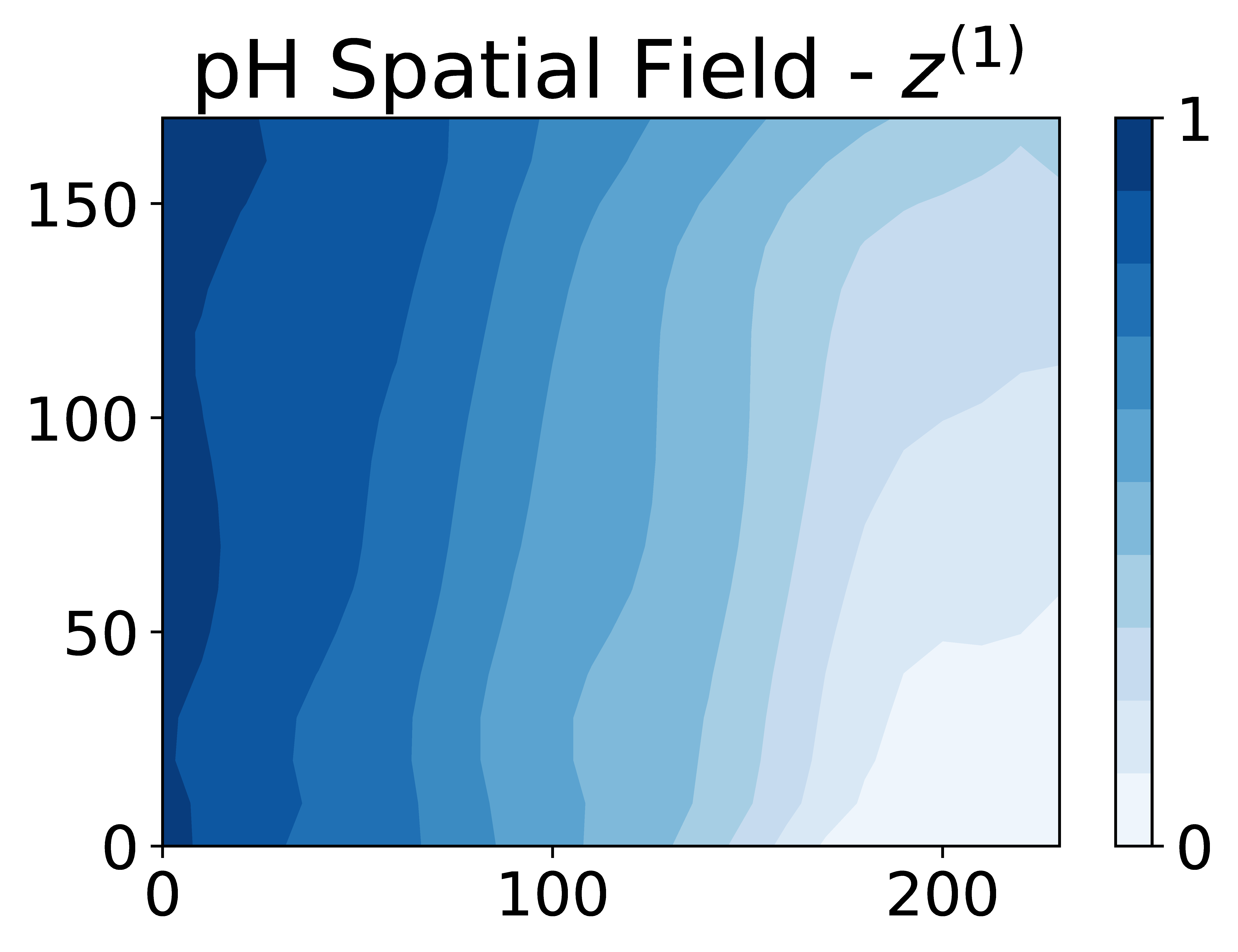}\hspace{-0.8em}
    \includegraphics[width = 0.33\columnwidth]{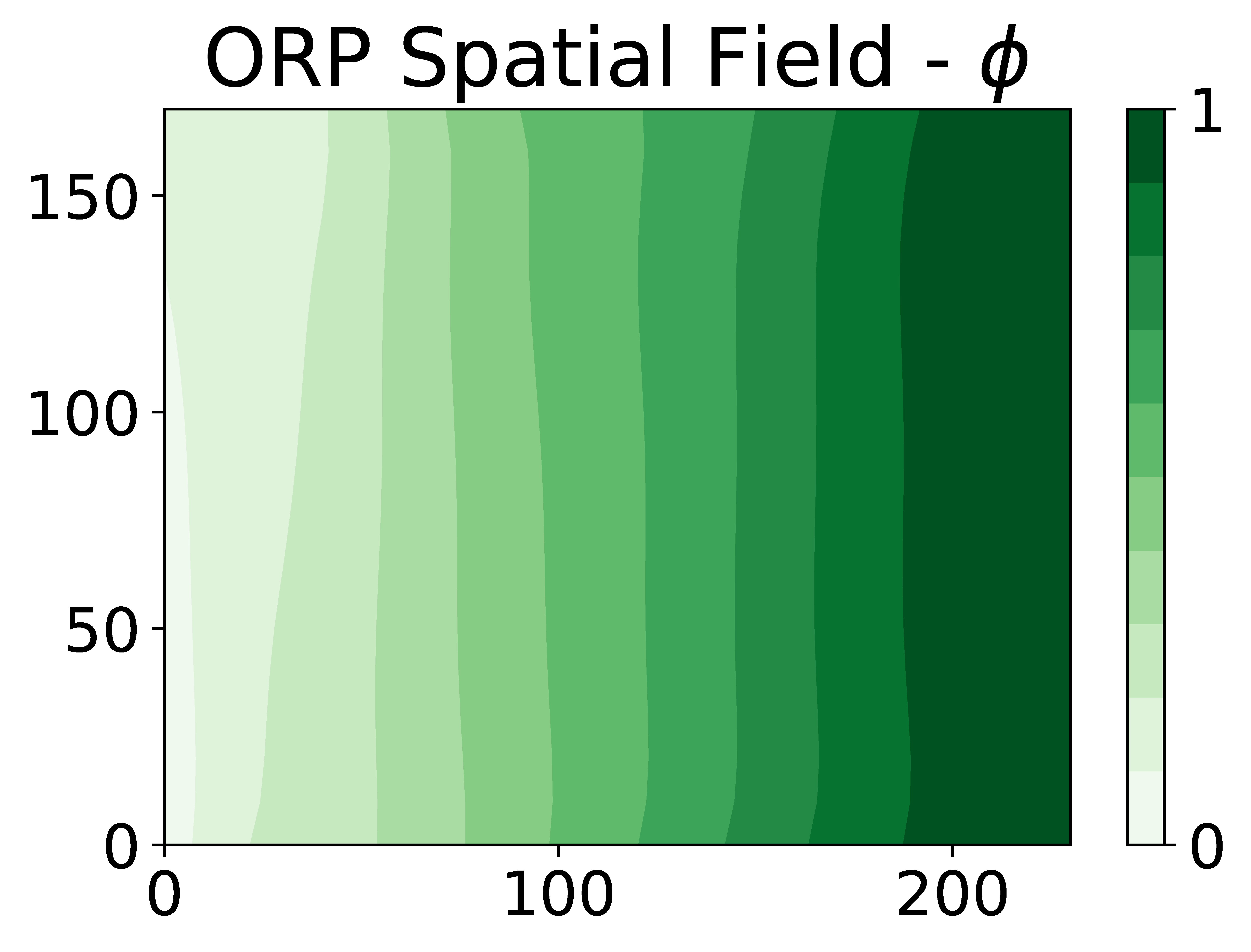}
    \caption{Our simulated environment includes two observable spatial fields: $z^{(0)}$ temperature (left), and $z^{(1)}$ pH (middle). These fields are unknown but observable:f $z^{(0)}$ can be measured by robot $r_0$ and $z^{(1)}$ can be measured by robot $r_1$. The simulated environment also includes a latent spatial field $\field$ of ORP values (right) which \textbf{cannot be measured by either robot.}}
    \label{fig:gt} \vspace{-1.5em}
\end{figure}

We present an instantiation of Algorithm~\ref{alg:planner} and investigate the impact of the correlations among the observable spatial fields described in Section~\ref{sec:interSensor}. This section presents simulated experiments where a team of robots estimate a latent phenomenon. We use real sensor data collected from a lake to create a simulated environment of two observable spatial fields and a latent phenomenon of interest. In the following subsections, we present a marine data set, we overview a simple myopic planner that instantiates Algorithm~\ref{alg:planner}, we describe the modeling of the correlations between the observable spatial fields, and we present results from simulated experiments. We show qualitatively that the learned correlations between the observable spatial fields $h_s$ improve the performance by causing the robot team to spread their samples out. Our quantitative evaluations show that these correlations improve the accuracy of the latent field estimate.

\subsection{Experimental Setup - Data and Simulated Environment} \label{sec:data}
We simulate a team of two robots in a marine setting. One of the robots is equipped with a temperature sensor and the other one with a pH sensor. Together, they are seeking information about oxidation reduction potential (ORP), the latent phenomenon of interest in these simulated experiments. Historical data temperature, pH, and ORP are needed to learn the correlation model $g$.

\textit{Lac Hertel Data.} The historical data used in our simulated experiments was collected on Lac Hertel in Mont Saint-Hilaire, Quebec. The full data set, detailed in the Appendix, is made publicly available. For this paper, we use temperature and pH data as our observable spatial fields, and the oxidation reduction potential (ORP) data as the latent phenomenon. ORP measures the ability of a lake or river to cleanse itself or break down waste products; ORP is aligned with the "knowledge" that might be sought in the environmental monitoring of a lake because it indicates whether drinking water is sanitary or if the water is suitability for anaerobic microbial processes. Additionally, we found that ORP, pH, and temperature had well-formed relations that could be modeled. Let the historical data $\mathcal{D} = (\mathcal{X}_{\mathcal{D}}, d_0, d_1, d_2)$ be a tuple of the sets of sample locations, $\mathcal{X}_{\mathcal{D}}$, the sets of measurements from two observable spatial fields (temperature:$d_0$ and pH:$d_1$) and the set of measurements from the latent spatial field (ORP:$d_2$). This historical data is used to learn the mapping from observations to the latent spatial field $g$ and the correlations $h_0$ and $h_1$.

\textit{Simulated Environment.}
To test Algorithm~\ref{alg:planner}, we create a simulated environment of observable and latent spatial fields. Fig.~\ref{fig:gt} shows the three spatial fields of our simulated test environment. All three fields are unknown. However, $z^(0)$ can be measured by robot $r_0$ and $z^(1)$ can be measured by robot $r_1$. The latent field $\field$ cannot be measured by the robot team; this spatial field is used only to evaluate the performance of our algorithm. We normalize all spatial fields to a scale from zero to one.\footnote{Normalizing the sensor data aids in learning correlation models.} To simulate our robots' motion in this simulated environment, we discretize the region into a grid, defining the set $\mathcal{X}$ of locations and corresponding ground truth sensor measurements $\mathcal{GT} = (gt_0, gt_1, gt_2)$.

\subsection{Myopic Planner}
For our simulated experiments, we assume a robot can move to any adjacent cell during a time step. For a given location $\pose_{t-1}^{(s)} \in \mathcal{X}$, the possible next locations $\pose_t^{(s)}$ in line~\ref{alg:line:acquisition} are the eight grid cells surrounding the current location (excluding cells outside of the boundary, i.e., $\notin \mathcal{X}$). 
Lines~\ref{alg:line:learner_init}~and~\ref{alg:line:update} create belief models of the observable spatial field $\Hat{\obs}$. At each time step $t$, the sensor measurements and respective sample locations are used to fit the belief model $\Hat{\obs}$ of the observable fields using Gaussian Processes. To create these beliefs, we assume known spatial correlations (covariance kernels), which is a common assumption because underlying physics informs these correlation (e.g., how temperature changes spatially). The belief model of the latent phenomenon, $\Hat{\field}$, is created/updated using the pre-learned mapping $g$ to predict the mean and variance of the latent spatial field.  The input to $g$ is $\Hat{\obs}$ and the output is $\Hat{\field}$. Importantly, we do not assume known spatial correlations of the latent variable; Mean and variance predictions are strictly a function of the observable spatial field belief model and the mapping $g$.

The mapping $g$ can be learned by various machine learning methods which capture the aleatoric uncertainty between observations and the latent spatial field \cite{hullermeier_aleatoric_2021}. Because we are only considering $S=2$ observable spatial fields, deep learning methods are not required. We use Gaussian Process Regression with inputs from $d_0$ and $d_1$ and targets of $d_2$ to train our mapping $g$. Then, to capture the uncertainty in the latent variable, we use samples at each location of our belief $\Hat{\obs}$ to capture the variance in the belief; we predict the value and variance of the latent spatial field for all samples; then, we calculate the mean and variance of the latent spatial field predictions at each point to create our latent belief $\Hat{\field}$.

To calculate the expected information gain in line~\ref{alg:line:acquisition}, we use the variance of the latent phenomenon estimate, $var(\Hat{\field})$. This is an approximation of Shannon entropy that is often used to reduce computational cost. The calculated Shannon entropy exactly, $H(\Hat{\field}) - H(\Hat{\field} | \pose)$, is computationally expensive because it requires that $\Hat{\field}$ be estimated for each action under consideration. When using a GP belief model, this requires a GP update for every action, for every robot, at each time step. The last piece of Algorithm~\ref{alg:planner} is the learned correlation model, $\interSensor_s$, which we detail in the next section.


\begin{figure}[tb]
  \begin{subfigure}{\columnwidth}
    \centering
    \includegraphics[width = 0.425\columnwidth]{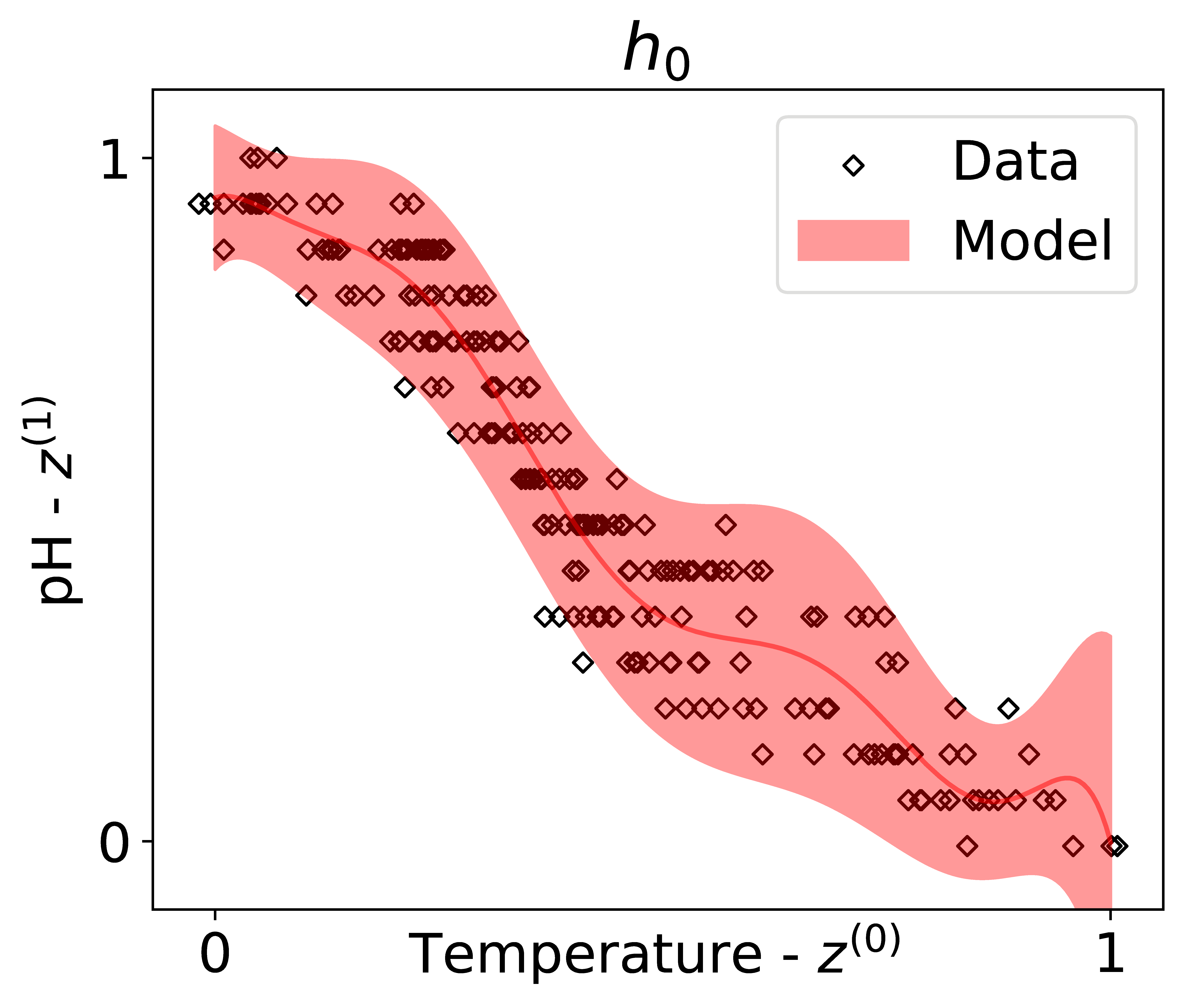}
    \includegraphics[width = 0.425\columnwidth]{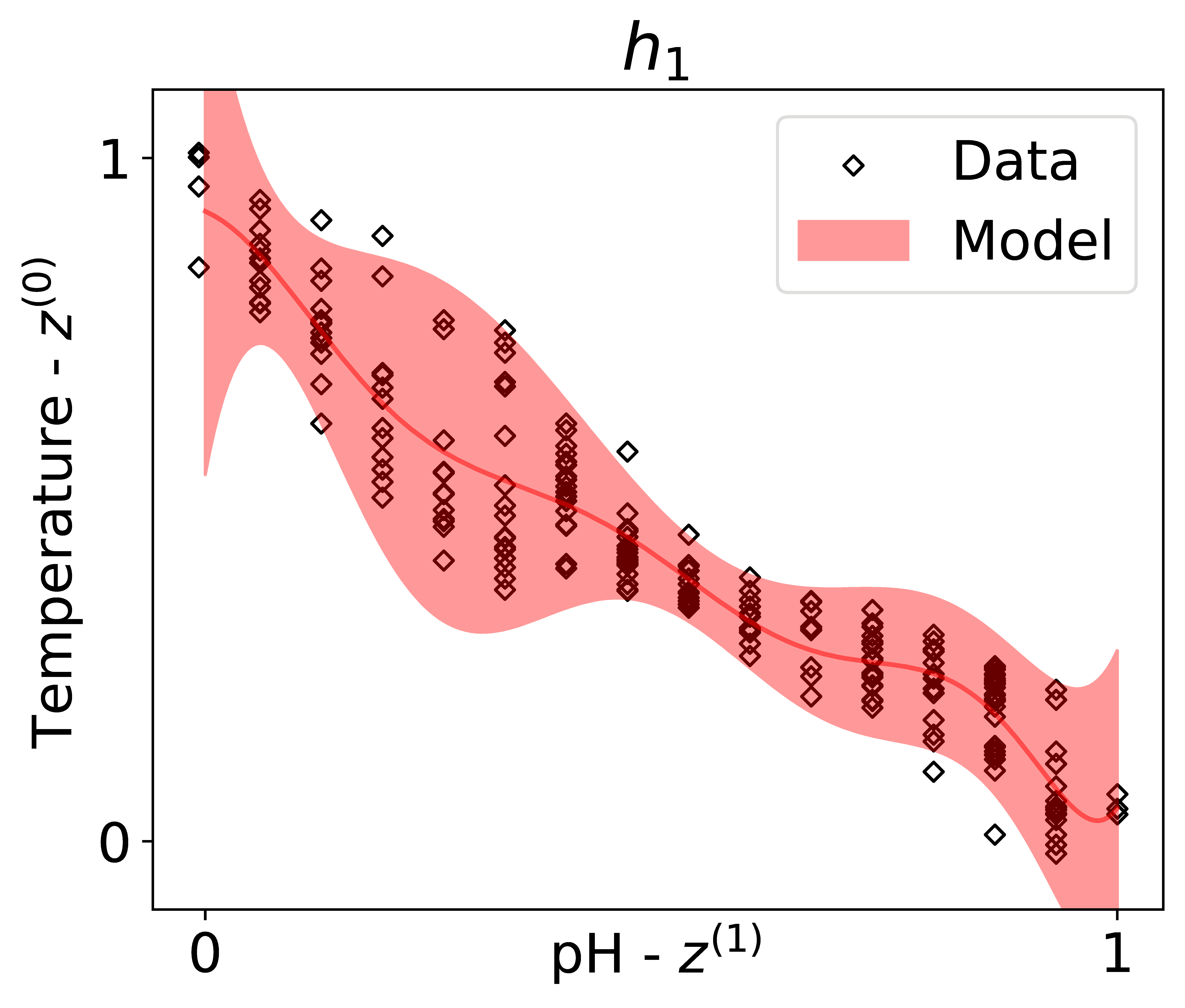}
    \caption{Correlation models} 
  \end{subfigure}
  
  \begin{subfigure}{\columnwidth}
    \centering
    \includegraphics[width = 0.55\columnwidth]{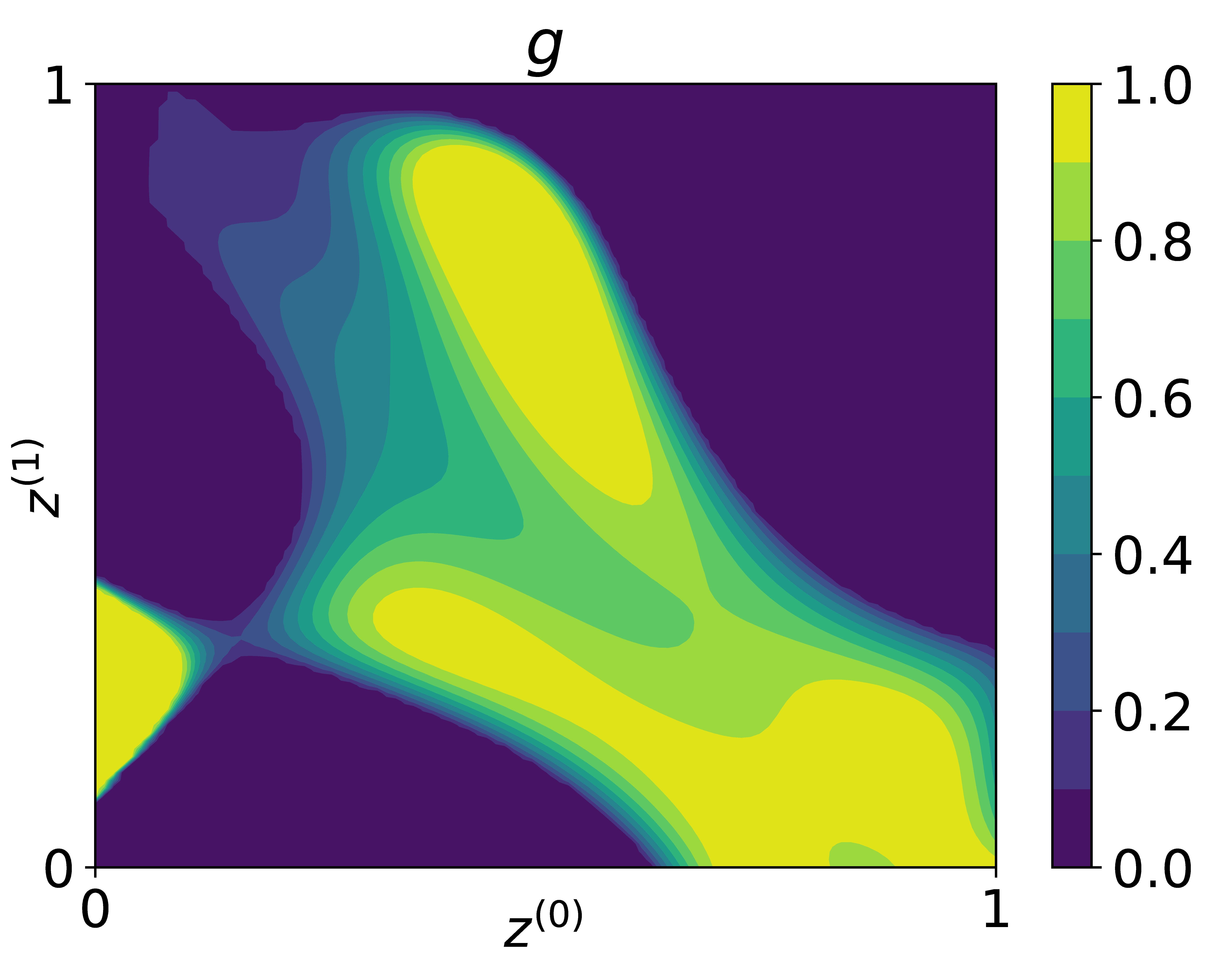}
    \caption{Mapping $g$ from $Z$ to $\phi$} 
  \end{subfigure}
    \caption{\textit{(a)} The learned correlation mappings $h_0$ from the pH spatial field to the temperature spatial field and $h_1$ from temperature to pH. The solid line shows the mean prediction and the bars show the variance prediction. \textit{(b)} The learned mapping $g$ from observations to the latent spatial field.}
    \label{fig:intersensor-data} \vspace{-1.5em}
\end{figure}

\subsection{Modeling Correlations among Observable Spatial Fields}
The core hypothesis of this work is that modeling correlations among observable spatial fields can improve information gathering. By capturing the aleatoric uncertainty in the relationship, robots can sense their own observable field and use the learned model to reduce the uncertainty of their teammates' observable spatial fields.

The goal here is to learn both $\interSensor_{0}$, the mapping from a sensor measurement in $\obsSingle[0]{}$ to a predicted value and variance in $\obsSingle[1]{}$, as well as $\interSensor_{1}$, the mapping from $\obsSingle[1]{}$ to $\obsSingle[0]{}$. Note that $\interSensor_{0}$ is not equivalent to the inverse of $\interSensor_{1}$ because both $\interSensor$ map to a variance, not just a predicted value. Learning such models requires supervised learning with input values and both target values and variance. However, the data discussed in Section~\ref{sec:data} does not include the target variance. For example, learning the mapping from temperature to pH requires the variance of pH for a given temperature data point, but we only have \textit{(temperature, pH)} value pairs, therefore, we transform the data. To learn $h_0$, the $d_0$ data binned and the the mean and standard deviation of the respective $d_1$ data are calculated. GP regression takes the $d_0$ bins as inputs and the $d_1$ means and standard deviations as targets. For learning $h_1$, the process is the same, flipping $d_0$ and $d_1$.

Fig.~\ref{fig:intersensor-data} shows the learned mappings $\interSensor_{0}$ and $\interSensor_{1}$ plotted over the respective historical data $d_0$ and $d_1$. These models are used in line~\ref{alg:line:infer}. When the robot equipped with the temperature sensor takes a measurement at the point $\pose^{(0)}_t$, the sensor observation $\obsSingle[0]{t}$ is used as the input to $\interSensor_{0}$ to update $\obsSingle[1]{t}(x)$, the belief of pH at point $\pose^{(0)}_t$.

\subsection{Evaluation} \label{sec:eval}
We evaluate the performance of Algorithm~\ref{alg:planner} with and without the correlation model, i.e., we omit line~\ref{alg:line:infer} in the latter. As the goal of Problem~\ref{prob:adaptive_latent} is to maximize the information with respect to the latent spatial field, the evaluation metric we use is the mean squared error between the ground truth of the latent phenomenon, $gt_2$, and the estimate $\Hat{\field}$, calculated using all points $x \in \mathcal{X}$ in the discretized space:

\begin{equation}\label{equ:mse}
MSE = \frac{1}{|\mathcal{X}|} \sum_{x \in \mathcal{X}} \left( gt_2(x) - \Hat{\field}(x)  \right)^2
\end{equation}

\begin{figure}[tb]
    \centering
    \includegraphics[width = 0.348\columnwidth]{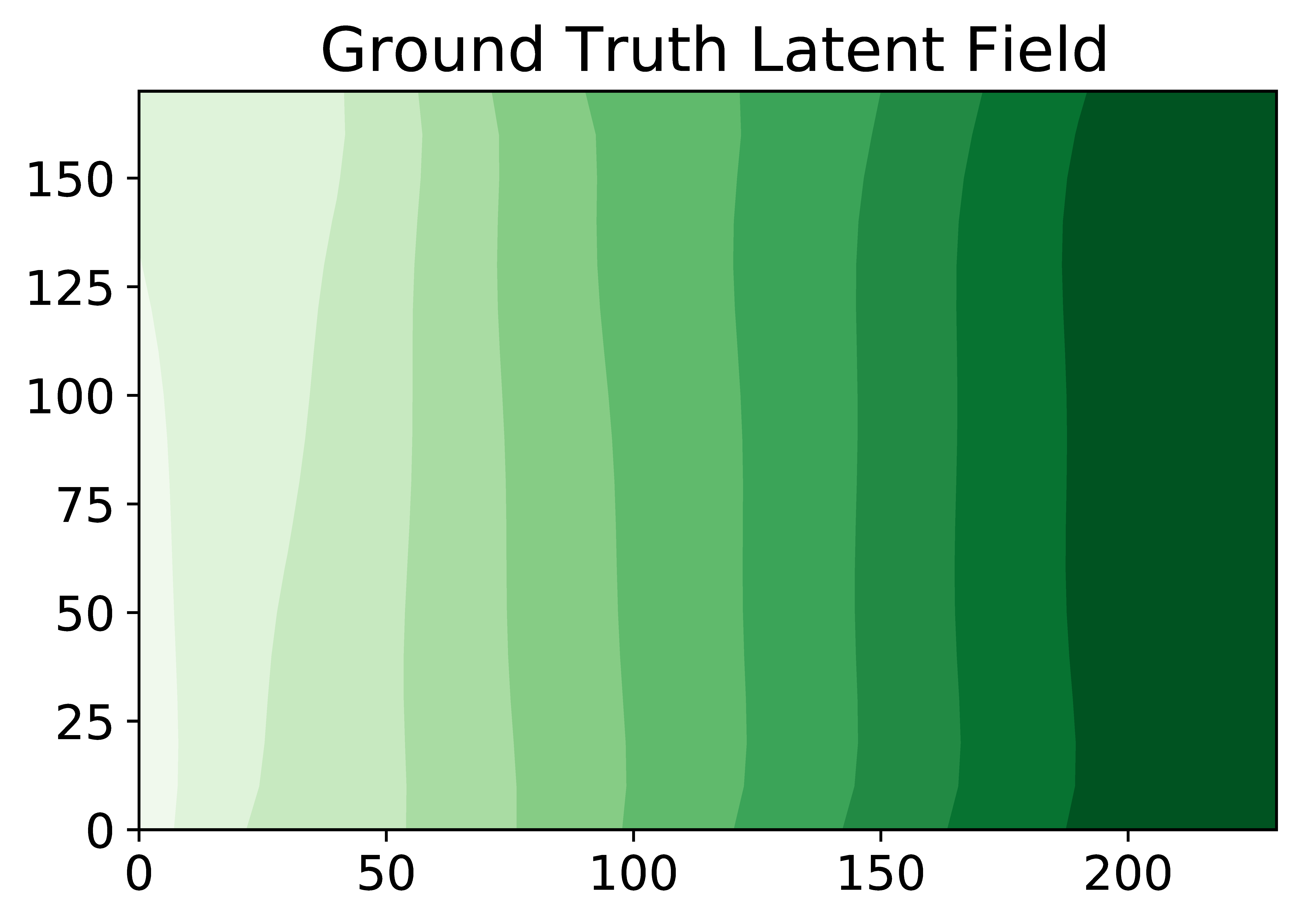}\hspace{-0.5em}
    \includegraphics[width = 0.321\columnwidth]{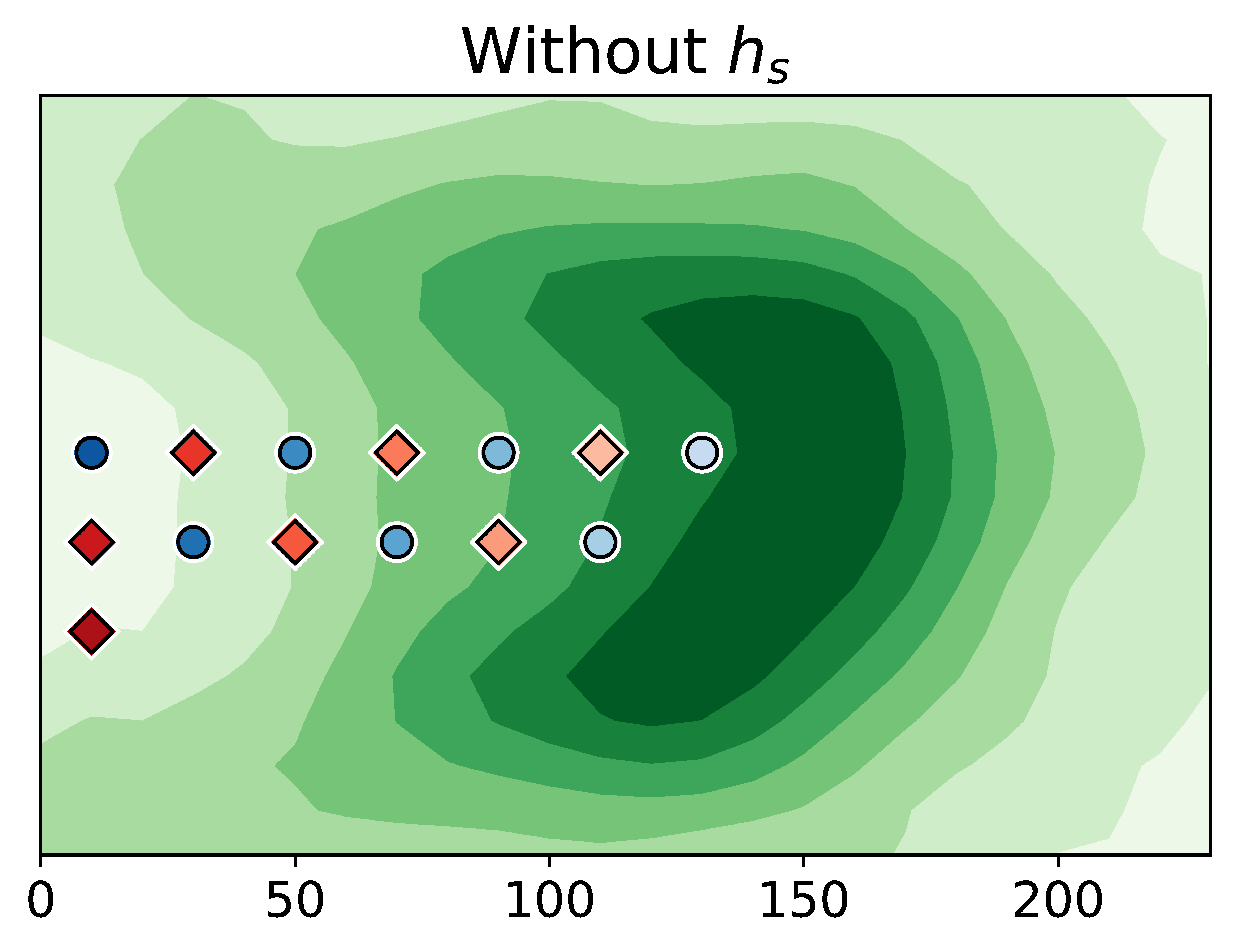}\hspace{-0.5em}
    \includegraphics[width = 0.321\columnwidth]{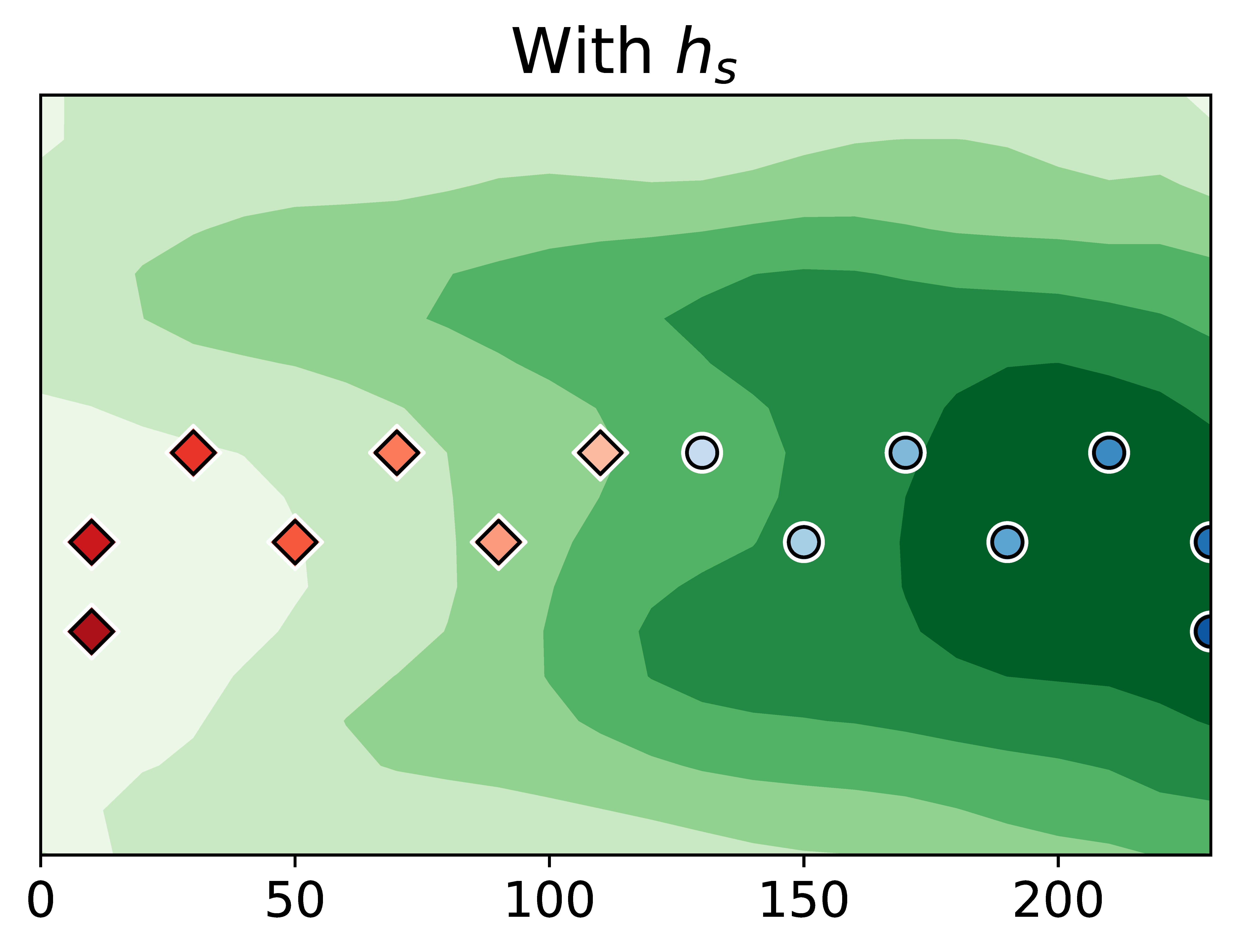}
    \caption{\textit{Left:} The ground truth of latent spatial field $\field$. \textit{Middle and right:} The belief model $\Hat{\field}$ of the latent phenomenon with robot measurements shown in red (square) and blue (circle) for the temperature and humidity robot, respectively; their sensing locations are indicated by color, with darker shades indicating later time steps. \textit{Middle:} The correlation models $h_s$ are not used by the robot team. \textit{Right:} When using the correlation models $h_s$, the team naturally spreads out their samples to maximize information gain.}
    \label{fig:qualitative}
\end{figure}

First, we run Algorithm~\ref{alg:planner} with the robots starting in the center of the region. The short time horizon mission in Fig.~\ref{fig:qualitative} shows a qualitative comparison of the algorithm's behavior with and without use of the correlation model. We see that when the correlation model is used, the robot team naturally spreads out to explore different parts of the region. Because of the correlations among the observable spatial fields, the robots are able to better estimate the expected information gain, finding that there is less information in the regions where other sensors have sampled because of the mutual information between the observable spaces.

\begin{figure}
    \centering
    \includegraphics[width = 0.7\columnwidth]{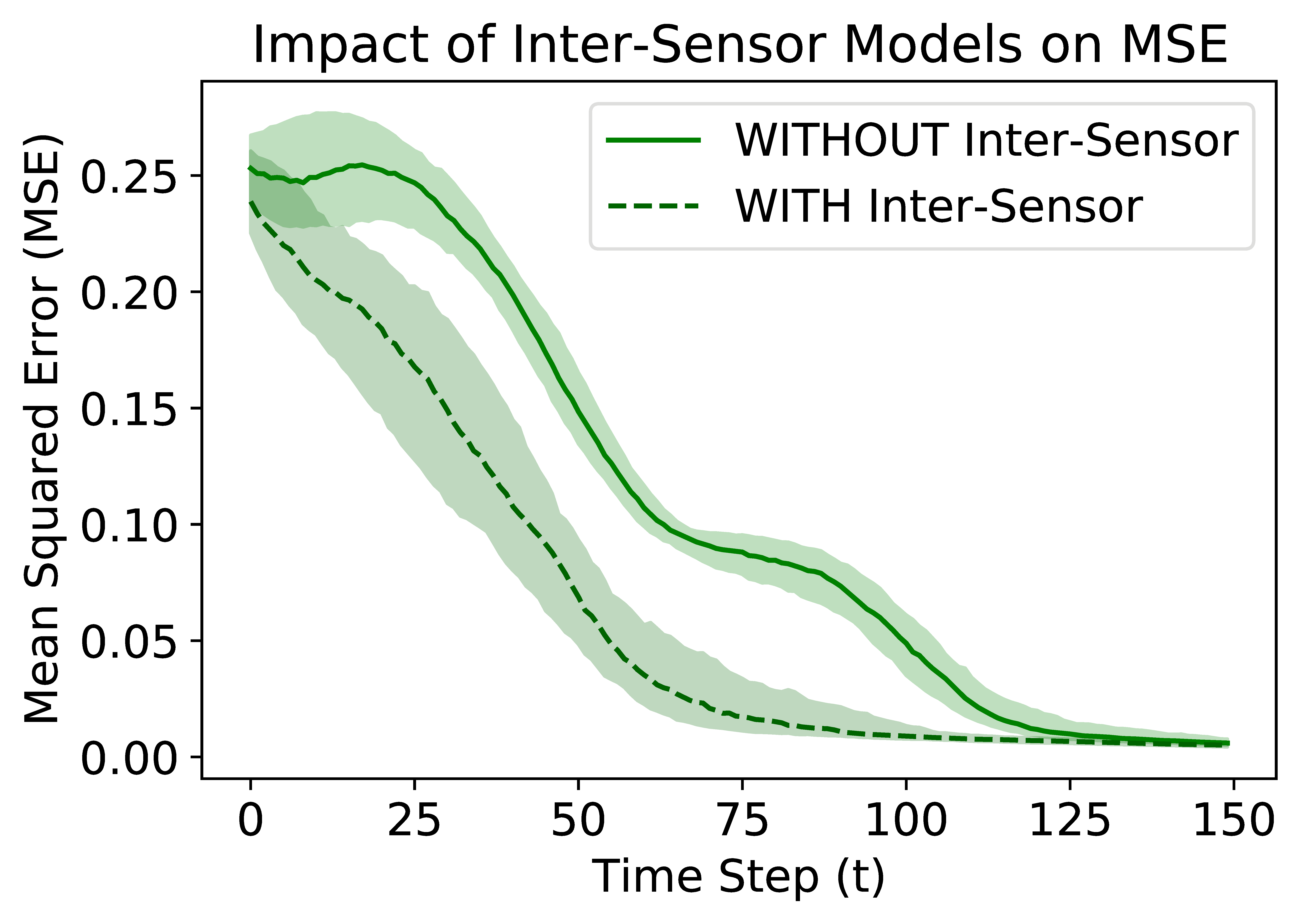}
    \caption{The mean squared error (MSE) of the belief model $\Hat{\field}$ shows that using learned correlations among the observable spatial fields improves the performance of the adaptive sampling approach. The solid and dashed lines show the median MSE at each time step for our instantiation of Algorithm without and with correlation model $h_s$, respectively. The shaded regions show the 25th to 75th percentile range.}
    \label{fig:mse1}
    \vspace{-1em}
\end{figure}

Next, we generate 200 trials by sampling sets of random starting locations from $\mathcal{X}$. We evaluate all time horizons up to $T=150$, reporting the MSE in Equation~\ref{equ:mse} for each time horizon. Fig.~\ref{fig:mse1} shows the median MSE among the 200 trials as well as the 25th to 75th percentile range for Algorithm~\ref{alg:planner}. The comparison of this algorithm with (dotted line) and without (solid line) the learned correlations illustrates that the learned correlations among the observable spatial fields improve the efficiency of the adaptive sampling algorithm.

In the earlier phase of the sampling survey, the improvement is small because the mutual information between sensor types can not yet be exploited and the randomness of starting locations has a large impact on performance. On the other hand, during the later phase of the sampling survey, the robot samples begin to cover the entirety of the space. Therefore, both methods converge close to zero MSE and zero variance. In practice, it is prohibitively expensive to sample the entire space (and robot sampling capacity or battery life may preclude this). Therefore, adaptive sampling methods often exist in the middle ground of time horizons, where enough samples are taken to provide good information about the region, but not so many samples that it becomes overly costly. In this realm, our method performs well, using the learned correlations among the observable spatial fields to maximize the amount of information gained about the latent phenomenon in a limited number of samples.

\section{Conclusion}\label{sec:Conculsion}
Real world robotic sampling methods often collect data to be analyzed offline by humans or machine learning algorithms to gain information about a latent phenomenon. However, the strength of robotic sampling is the ability for real-time adaptation. To bridge this gap, we present the ASLaP problem, where robot sampling decisions are made with respect to the information gain of a latent spatial field. We propose an informed path planning approach, ASLaP-HR to this problem using a team of heterogeneous robots.

Each robot on the team has a distinct sensor type that can measure an observable spatial field. Collectively, the measurements made by the robot team inform an estimate of the latent spatial field by using a machine learning model trained on historical data. Importantly, this learned model is used online by the robots to calculate the expected information gain of their actions with respect to the latent phenomenon. \textbf{Our method is among the firsts to explore adaptive sampling of a latent phenomenon with a team of robots using a learned model for online decision making. Additionally, our adaptive sampling strategy uses learned correlation models among heterogeneous robots to improve the performance of the team.}

We recognize that different sensor types in the real world are not independent but rather have mutual information that is a complex unknown function of their measurements. Therefore, we hypothesize that these heterogeneous robots can improve their information gathering by modeling this mutual information. We use historical sensor data to learn correlations among the observable spatial fields, and we use these learned models in our informed path planning approach. To test the impact of these correlation models, we create a simulated environment of two observable spatial fields (temp. and pH) and a latent spatial field (ORP). We run our algorithm in this simulated environment and calculate the mean squared error between the latent spatial field estimate returned by our algorithm and the ground truth latent spatial field. We aggregate our results across hundreds of randomized starting locations, showing that the learned correlations improve the efficiency of our ASLaP-HR approach. 

\section{Future Directions}\label{sec:Discuss}
As seen in Section~\ref{Sec:Prelim}, there are myriad approaches to adaptive sampling and information gain with robot teams. The many dimensions of these approaches can be incorporated into our latent sampling framework; this includes complex motion primitives, long planning horizons, time-varying phenomena, objectives such as seeking the maximum of the latent spatial field, and uncertainty in sensors or motion, to name a few. While these open problems are shared among the adaptive sampling community, latent adaptive sampling brings new open problems. For example, new learning methods are needed for the mapping $g$ to better capture aleatoric uncertainty in the relationship between observable and latent spatial fields; current supervised learning methods focus on aleatoric uncertainty in classification, with less work on regression. An important next step that this paper enables is further investigation of the correlation models. Specifically, these systems would benefit from online learning or continual learning of these models. Such approaches would require the robots to navigate the exploration-exploitation trade-off with respect to the correlation models. In this work, we assume full communication between the robots. One near future extension of this work will be to consider varying communication ranges and intermittent communication \cite{manjanna_scalable_2021}.

\appendix
We make public the data set used in our second set of experiments. Data consist of water quality measurements collected at Lac Hertel in Mont Saint-Hilaire, Quebec. This data is collected using a multiparameter Sonde, Exo1 from YSI. The sensor was mounted on an autonomous surface vehicle to collect these data measurements. The data includes geotagged and timestamped measurements of 10 water quality parameters measured using four sensors: temperature, conductance, salinity, pH, oxidation reduction potential (ORP), depth, two turbidity measures and two optical dissolved oxygen measures. We are making this data public through the following Github repository: \url{https://github.com/SandeepManjanna/water_quality_dataset}

\begin{figure}[H]
    \centering
    \begin{subfigure}{0.48\columnwidth}
    \includegraphics[width = \columnwidth]{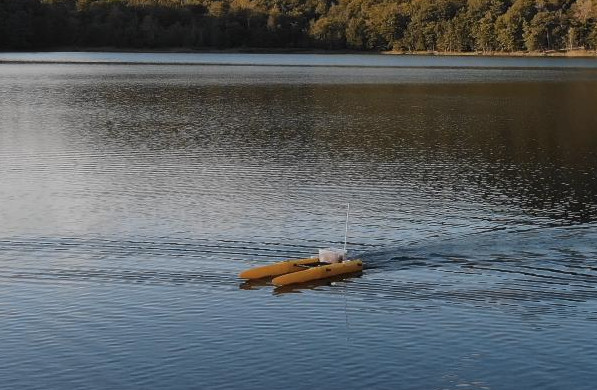}
    \caption{ASV collecting data at Lac Hertel} 
    \label{fig:boat}
  \end{subfigure}%
  \hfill
  \begin{subfigure}{0.42\columnwidth}
    \centering
    \includegraphics[width = \columnwidth]{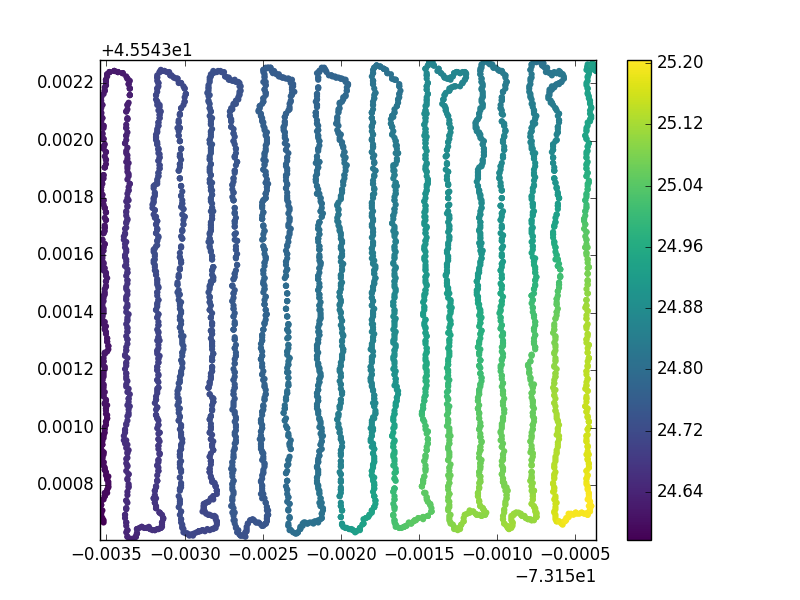}
    \caption{Temperature measurements} 
    \label{fig:temperature}
  \end{subfigure}%
    \caption{}
    \label{fig:appendix}
    \vspace{-1em}
\end{figure}

\bibliographystyle{IEEEtran}
\bibliography{references}

\begin{thebibliography}{10}
\providecommand{\url}[1]{#1}
\csname url@samestyle\endcsname
\providecommand{\newblock}{\relax}
\providecommand{\bibinfo}[2]{#2}
\providecommand{\BIBentrySTDinterwordspacing}{\spaceskip=0pt\relax}
\providecommand{\BIBentryALTinterwordstretchfactor}{4}
\providecommand{\BIBentryALTinterwordspacing}{\spaceskip=\fontdimen2\font plus
\BIBentryALTinterwordstretchfactor\fontdimen3\font minus
  \fontdimen4\font\relax}
\providecommand{\BIBforeignlanguage}[2]{{%
\expandafter\ifx\csname l@#1\endcsname\relax
\typeout{** WARNING: IEEEtran.bst: No hyphenation pattern has been}%
\typeout{** loaded for the language `#1'. Using the pattern for}%
\typeout{** the default language instead.}%
\else
\language=\csname l@#1\endcsname
\fi
#2}}
\providecommand{\BIBdecl}{\relax}
\BIBdecl

\bibitem{dunbabin_robots_2012}
M.~Dunbabin and L.~Marques, ``Robots for environmental monitoring:
  {Significant} advancements and applications,'' \emph{IEEE Robotics \&
  Automation Magazine}, vol.~19, no.~1, pp. 24--39, 2012, publisher: IEEE.

\bibitem{manjanna_scalable_2021}
S.~Manjanna, M.~A. Hsieh, and G.~Dudek, ``Scalable {Multi}-{Robot} {System} for
  {Non}-myopic {Spatial} {Sampling},'' \emph{arXiv preprint arXiv:2105.10018},
  2021.

\bibitem{barker_relationships_1976}
K.~Barker and T.~H. Olthof, ``Relationships between nematode population
  densities and crop responses,'' \emph{Annual Review of Phytopathology},
  vol.~14, no.~1, pp. 327--353, 1976, publisher: Annual Reviews 4139 El Camino
  Way, PO Box 10139, Palo Alto, CA 94303-0139, USA.

\bibitem{quattrini_li_multi-robot_2020}
A.~Quattrini~Li, P.~K. Penumarthi, J.~Banfi, N.~Basilico, J.~M. O’Kane,
  I.~Rekleitis, S.~Nelakuditi, and F.~Amigoni, ``Multi-robot online sensing
  strategies for the construction of communication maps,'' \emph{Autonomous
  Robots}, vol.~44, no.~3, pp. 299--319, 2020, publisher: Springer.

\bibitem{almadhoun_survey_2019}
R.~Almadhoun, T.~Taha, L.~Seneviratne, and Y.~Zweiri, ``A survey on multi-robot
  coverage path planning for model reconstruction and mapping,'' \emph{SN
  Applied Sciences}, vol.~1, no.~8, pp. 1--24, 2019, publisher: Springer.

\bibitem{manjanna_heterogeneous_2018}
S.~Manjanna, A.~Q. Li, R.~N. Smith, I.~Rekleitis, and G.~Dudek, ``Heterogeneous
  {Multi}-{Robot} {System} for {Exploration} and {Strategic} {Water}
  {Sampling},'' in \emph{2018 {IEEE} {Intl.} {Conf.} on {Robotics} and
  {Automation} ({ICRA})}, 2018, pp. 4873--4880.

\bibitem{arora_multi-modal_2019}
A.~Arora, P.~M. Furlong, R.~Fitch, S.~Sukkarieh, and T.~Fong, ``Multi-modal
  active perception for information gathering in science missions,''
  \emph{Autonomous Robots}, vol.~43, no.~7, pp. 1827--1853, 2019, publisher:
  Springer.

\bibitem{salam_heterogeneous_2021}
T.~Salam and M.~A. Hsieh, ``Heterogeneous robot teams for modeling and
  prediction of multiscale environmental processes,'' \emph{arXiv preprint
  arXiv:2103.10383}, 2021.

\bibitem{manderson_heterogeneous_2019}
T.~Manderson, S.~Manjanna, and G.~Dudek, ``Heterogeneous robot teams for
  informative sampling,'' \emph{arXiv preprint arXiv:1906.07208}, 2019.

\bibitem{shi_adaptive_2020}
Y.~Shi, N.~Wang, J.~Zheng, Y.~Zhang, S.~Yi, W.~Luo, and K.~Sycara, ``Adaptive
  informative sampling with environment partitioning for heterogeneous
  multi-robot systems,'' in \emph{2020 {IEEE}/{RSJ} {Intl.} {Conf.} on
  {Intelligent} {Robots} and {Systems} ({IROS})}.\hskip 1em plus 0.5em minus
  0.4em\relax IEEE, 2020, pp. 11\,718--11\,723.

\bibitem{chlingaryan_machine_2018}
A.~Chlingaryan, S.~Sukkarieh, and B.~Whelan, ``Machine learning approaches for
  crop yield prediction and nitrogen status estimation in precision
  agriculture: {A} review,'' \emph{Computers and electronics in agriculture},
  vol. 151, pp. 61--69, 2018, publisher: Elsevier.

\bibitem{marchant_sequential_2014}
R.~Marchant, F.~Ramos, S.~Sanner, and {others}, ``Sequential {Bayesian}
  optimisation for spatial-temporal monitoring.'' in \emph{{UAI}}, 2014, pp.
  553--562.

\bibitem{oliveira_bayesian_2019}
R.~Oliveira, L.~Ott, and F.~Ramos, ``Bayesian optimisation under uncertain
  inputs,'' in \emph{The 22nd Intl. conf. on artificial intelligence and
  statistics}.\hskip 1em plus 0.5em minus 0.4em\relax PMLR, 2019, pp.
  1177--1184.

\bibitem{tan_gaussian_2018}
Y.~T. Tan, A.~Kunapareddy, and M.~Kobilarov, ``Gaussian process adaptive
  sampling using the cross-entropy method for environmental sensing and
  monitoring,'' in \emph{2018 {IEEE} {Intl.} {Conf.} on {Robotics} and
  {Automation} ({ICRA})}.\hskip 1em plus 0.5em minus 0.4em\relax IEEE, 2018,
  pp. 6220--6227.

\bibitem{yamauchi_frontier-based_1998}
B.~Yamauchi, ``Frontier-based exploration using multiple robots,'' in
  \emph{Proceedings of the second intl. Conf. on {Autonomous} agents}, 1998,
  pp. 47--53.

\bibitem{lluvia_active_2021}
I.~Lluvia, E.~Lazkano, and A.~Ansuategi, ``Active mapping and robot
  exploration: {A} survey,'' \emph{Sensors}, vol.~21, no.~7, p. 2445, 2021,
  publisher: Multidisciplinary Digital Publishing Institute.

\bibitem{popovic_informative_2020}
M.~Popović, T.~Vidal-Calleja, J.~J. Chung, J.~Nieto, and R.~Siegwart,
  ``Informative path planning for active field mapping under localization
  uncertainty,'' in \emph{2020 {IEEE} {Intl.} {Conf.} on {Robotics} and
  {Automation} ({ICRA})}.\hskip 1em plus 0.5em minus 0.4em\relax IEEE, 2020,
  pp. 10\,751--10\,757.

\bibitem{tzes_technical_2021}
M.~Tzes, Y.~Kantaros, and G.~J. Pappas, ``Distributed sampling-based planning
  for non-myopic active information gathering,'' in \emph{2021 IEEE/RSJ Intl.
  Conf. on Intelligent Robots and Systems (IROS)}.\hskip 1em plus 0.5em minus
  0.4em\relax IEEE, 2021, pp. 5872--5877.

\bibitem{kantaros_sampling-based_2021}
Y.~Kantaros, B.~Schlotfeldt, N.~Atanasov, and G.~J. Pappas, ``Sampling-based
  planning for non-myopic multi-robot information gathering,'' \emph{Autonomous
  Robots}, vol.~45, no.~7, pp. 1029--1046, 2021, publisher: Springer.

\bibitem{salam_adaptive_2019}
T.~Salam and M.~A. Hsieh, ``Adaptive {Sampling} and {Reduced}-{Order}
  {Modeling} of {Dynamic} {Processes} by {Robot} {Teams},'' \emph{IEEE Robotics
  and Automation Letters}, vol.~4, no.~2, pp. 477--484, 2019.

\bibitem{fung_coordinating_2019}
N.~Fung, J.~Rogers, C.~Nieto, H.~I. Christensen, S.~Kemna, and G.~Sukhatme,
  ``Coordinating multi-robot systems through environment partitioning for
  adaptive informative sampling,'' in \emph{2019 {Intl.} {Conf.} on {Robotics}
  and {Automation} ({ICRA})}.\hskip 1em plus 0.5em minus 0.4em\relax IEEE,
  2019, pp. 3231--3237.

\bibitem{bai_information-driven_2021}
S.~Bai, T.~Shan, F.~Chen, L.~Liu, and B.~Englot, ``Information-{Driven} {Path}
  {Planning},'' \emph{Current Robotics Reports}, vol.~2, no.~2, pp. 177--188,
  2021, publisher: Springer.

\bibitem{low_information-theoretic_2009}
K.~H. Low, J.~Dolan, and P.~Khosla, ``Information-theoretic approach to
  efficient adaptive path planning for mobile robotic environmental sensing,''
  in \emph{Proceedings of the {Intl.} conf. on automated planning and
  scheduling}, vol.~19, 2009, pp. 233--240.

\bibitem{singh_efficient_2009}
A.~Singh, A.~Krause, C.~Guestrin, and W.~J. Kaiser, ``Efficient informative
  sensing using multiple robots,'' \emph{Journal of Artificial Intelligence
  Research}, vol.~34, pp. 707--755, 2009.

\bibitem{low_adaptive_2008}
K.~H. Low, J.~M. Dolan, and P.~Khosla, ``Adaptive multi-robot wide-area
  exploration and mapping,'' in \emph{Proceedings of the 7th {Intl.} {Joint}
  {Conf.} on {Autonomous} agents and {Multiagent} systems-{Volume} 1}, 2008,
  pp. 23--30.

\bibitem{manjanna_policy_2018}
S.~Manjanna, H.~V. Hoof, and G.~Dudek, ``Policy search on aggregated state
  space for active sampling,'' in \emph{Intl. {Symposium} on {Experimental}
  {Robotics}}.\hskip 1em plus 0.5em minus 0.4em\relax Springer, 2018, pp.
  211--221.

\bibitem{chand_mapping_2013}
P.~Chand and D.~A. Carnegie, ``Mapping and exploration in a hierarchical
  heterogeneous multi-robot system using limited capability robots,''
  \emph{Robotics and autonomous Systems}, vol.~61, no.~6, pp. 565--579, 2013,
  publisher: Elsevier.

\bibitem{cai_non-monotone_2021}
X.~Cai, B.~Schlotfeldt, K.~Khosoussi, N.~Atanasov, G.~J. Pappas, and J.~P. How,
  ``Non-{Monotone} {Energy}-{Aware} {Information} {Gathering} for
  {Heterogeneous} {Robot} {Teams},'' in \emph{2021 {IEEE} {Intl.} {Conf.} on
  {Robotics} and {Automation} ({ICRA})}.\hskip 1em plus 0.5em minus 0.4em\relax
  IEEE, 2021, pp. 8859--8865.

\bibitem{newaz_multi-robot_2021}
A.~A.~R. Newaz, T.~Alam, J.~Mondello, J.~Johnson, and L.~Bobadilla,
  ``Multi-{Robot} {Information} {Gathering} {Subject} to {Resource}
  {Constraints},'' in \emph{2021 30th {IEEE} {Intl.} {Conf.} on {Robot} \&
  {Human} {Interactive} {Communication} ({RO}-{MAN})}.\hskip 1em plus 0.5em
  minus 0.4em\relax IEEE, 2021, pp. 1--6.

\bibitem{notomista_multi-robot_2022}
G.~Notomista, C.~Pacchierotti, and P.~R. Giordano, ``Multi-robot persistent
  environmental monitoring based on constraint-driven execution of learned
  robot tasks,'' in \emph{{ICRA} 2022-{IEEE} {Intl.} {Conf.} on {Robotics} and
  {Automation}}, 2022.

\bibitem{arora_randomized_2017}
S.~Arora and S.~Scherer, ``Randomized algorithm for informative path planning
  with budget constraints,'' in \emph{2017 {IEEE} {Intl.} {Conf.} on {Robotics}
  and {Automation} ({ICRA})}.\hskip 1em plus 0.5em minus 0.4em\relax IEEE,
  2017, pp. 4997--5004.

\bibitem{tiwari_multi-robot_2019}
K.~Tiwari and N.~Y. Chong, \emph{Multi-{Robot} {Exploration} for
  {Environmental} {Monitoring}: {The} {Resource} {Constrained}
  {Perspective}}.\hskip 1em plus 0.5em minus 0.4em\relax Academic Press, 2019.

\bibitem{chen_multi-objective_2019}
W.~Chen and L.~Liu, ``Multi-objective and model-predictive tree search for
  spatiotemporal informative planning,'' in \emph{2019 {IEEE} 58th {Conf.} on
  {Decision} and {Control} ({CDC})}.\hskip 1em plus 0.5em minus 0.4em\relax
  IEEE, 2019, pp. 5716--5722.

\bibitem{hollinger_sampling-based_2014}
G.~A. Hollinger and G.~S. Sukhatme, ``Sampling-based robotic information
  gathering algorithms,'' \emph{The Intl. Journal of Robotics Research},
  vol.~33, no.~9, pp. 1271--1287, 2014, publisher: SAGE Publications Sage UK:
  London, England.

\bibitem{hullermeier_aleatoric_2021}
E.~Hüllermeier and W.~Waegeman, ``Aleatoric and epistemic uncertainty in
  machine learning: {An} introduction to concepts and methods,'' \emph{Machine
  Learning}, vol. 110, no.~3, pp. 457--506, 2021, publisher: Springer.

\bibitem{krause_near-optimal_2008}
A.~Krause, A.~Singh, and C.~Guestrin, ``Near-optimal sensor placements in
  {Gaussian} processes: {Theory}, efficient algorithms and empirical studies.''
  \emph{Journal of Machine Learning Research}, vol.~9, no.~2, 2008.

\end{thebibliography}
\end{document}